\documentclass[twocolumn,english,10pt,journal,cspaper,compsoc]{IEEEtran}

\usepackage{blindtext}
\usepackage[pdftex]{graphicx}
\usepackage{amsfonts}
\usepackage{amsmath}
\usepackage{array}
\usepackage{mdwmath}
\usepackage{mdwtab}
\usepackage{eqparbox}
\usepackage{fixltx2e}
\usepackage{stfloats}
\usepackage{url}
\usepackage{diagbox}
\usepackage{verbatim}
\usepackage{booktabs}
\usepackage{multirow}
\usepackage{bm}
\usepackage{mathtools}
\usepackage{breqn}

\usepackage{float}
\usepackage{tcolorbox}
\usepackage{subcaption}
\usepackage{todonotes}
\usepackage{pgfmath}

\usepackage{threeparttable}
\usepackage{flushend}

\ifCLASSOPTIONcompsoc
  \usepackage[nocompress]{cite}
\else
  \usepackage{cite}
\fi

\newcommand{\etal}{et al.}

\DeclareMathOperator*{\argmax}{arg\,max}

\usepackage{algorithmic}

\floatstyle{ruled}
\newfloat{algorithm}{tbp}{loa}
\providecommand{\algorithmname}{Algorithm}
\floatname{algorithm}{\protect\algorithmname}

\begin{document}

\title{On the Capacity of Face Representation}

\author{Sixue~Gong,~\IEEEmembership{Student Member,~IEEE,}
        Vishnu~Naresh~Boddeti,~\IEEEmembership{Member, IEEE}
        and~Anil~K.~Jain,~\IEEEmembership{Life~Fellow,~IEEE}
\thanks{The authors are with the Department of Computer Science and Engineering, Michigan State University, East Lansing, MI, 48824 USA e-mail: (vishnu@msu.edu).}}


\IEEEtitleabstractindextext{%
\begin{abstract}
	In this paper we address the following question, \textit{given a face representation, how many identities can it resolve?} In other words, \textit{what is the capacity of the face representation?} A scientific basis for estimating the capacity of a given face representation will not only benefit the evaluation and comparison of different representation methods, but will also establish an upper bound on the scalability of an automatic face recognition system. We cast the face capacity problem in terms of packing bounds on a low-dimensional manifold embedded within a deep representation space. By explicitly accounting for the manifold structure of the representation as well two different sources of representational noise: \emph{epistemic} (model) uncertainty and \emph{aleatoric} (data) variability, our approach is able to estimate the capacity of a given face representation. To demonstrate the efficacy of our approach, we estimate the capacity of two deep neural network based face representations, namely 128-dimensional FaceNet and 512-dimensional SphereFace. Numerical experiments on unconstrained faces (IJB-C) provides a capacity upper bound of $2.7\times10^4$ for FaceNet and $8.4\times10^4$ for SphereFace representation at a false acceptance rate (FAR) of 1\%. As expected, capacity reduces drastically at lower FARs. The capacity at FAR of 0.1\% and 0.001\% is $2.2\times10^3$ and $1.6\times10^{1}$, respectively for FaceNet and $3.6\times10^3$ and $6.0\times10^0$, respectively for SphereFace.
\end{abstract}

}

\maketitle
\IEEEdisplaynontitleabstractindextext
\IEEEpeerreviewmaketitle


\section{Introduction}

Face recognition has witnessed rapid progress and wide applicability in a variety of practical applications: social media, surveillance systems and law enforcement to name a few. Fueled by copious amounts of data, ever growing computational resources and algorithmic developments, current state-of-the-art face recognition systems are believed to surpass human capability in certain scenarios \cite{lu2015surpassing}. Despite this tremendous progress, a crucial question still remains unaddressed, \emph{what is the capacity of a given face representation?} Tackling this question is the central aim of this paper.

Consider the following scenario: we would like to deploy a face recognition system with representation $M$ in a target application that requires a maximum false acceptance rate (FAR) of $q\%$. As we continue to add subjects to the gallery, it is known, empirically, that the face recognition accuracy starts decreasing. This is primarily due to the fact that with more subjects and diverse viewpoints, the representations of the classes will no longer be disjoint. In other words, the face recognition system based on representation $M$ can no longer completely resolve all of the users within the $q\%$ FAR. We define the maximal number of users at which the face representation reaches this limit as the capacity\footnote{This is different from the notion of capacity of a space of functions as measured by its VC dimension.} of the representation. Our contribution, in this paper, is to determine the capacity in an objective manner without the need for empirical evaluation.
\begin{figure*}[!ht]
    \centering
    \includegraphics[width=0.95\textwidth]{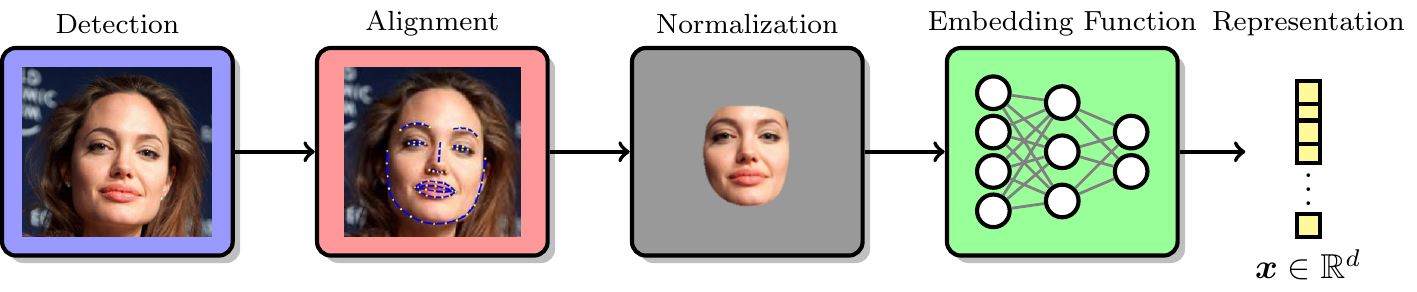}
    \caption{A typical \textbf{face representation pipeline} comprises of face detection, alignment, normalization and representation or feature extraction. While each of these components affect the capacity of the representation, in this paper, we focus on the capacity of the embedding function that maps a high-dimensional normalized face image to a $d$-dimensional vector representation.}
    \label{fig:pipeline}
\end{figure*}
\begin{figure}[!ht]
\centering
\includegraphics[width=0.45\textwidth]{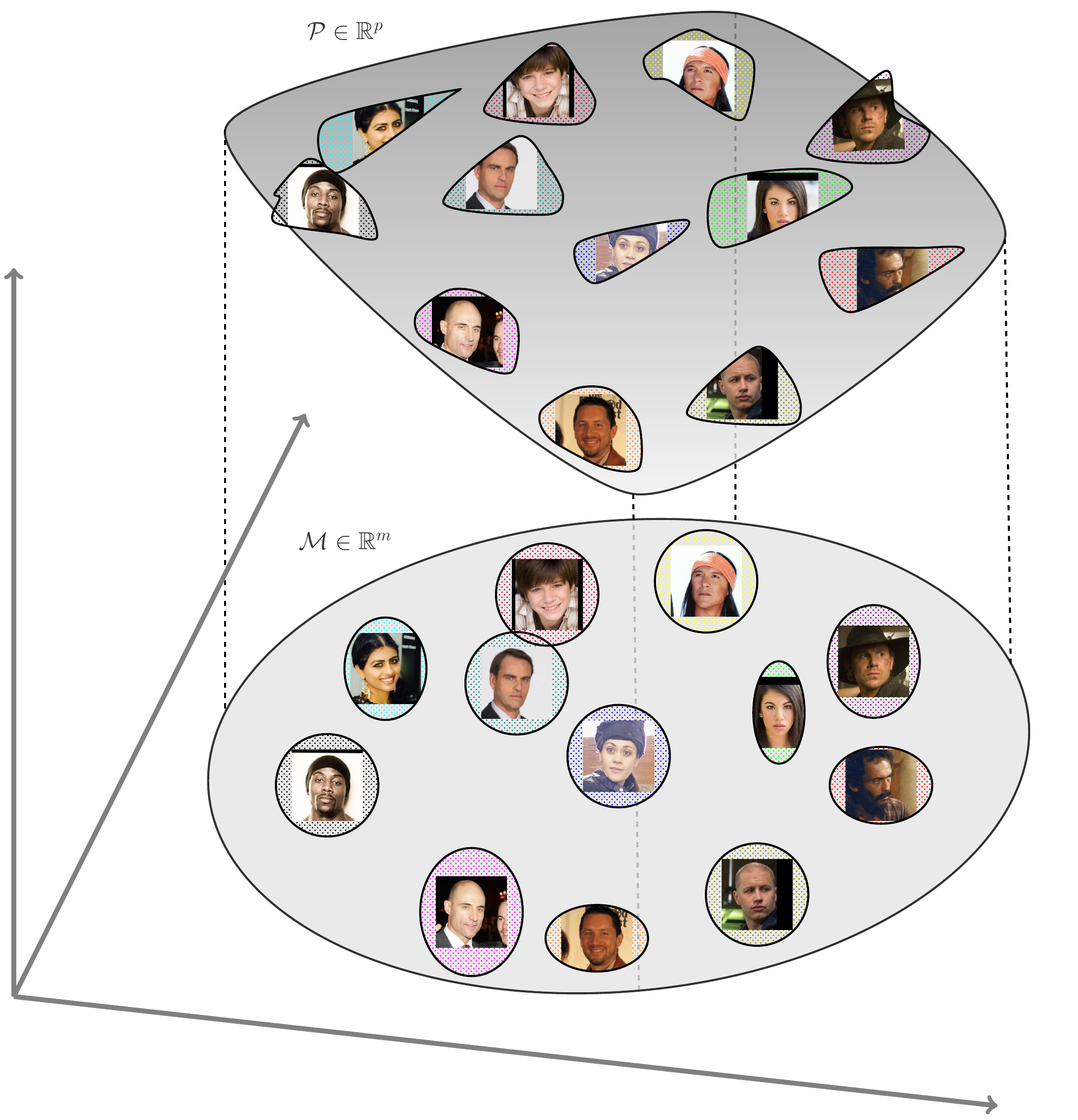}
\caption{An illustration of the geometrical structure of our capacity estimation problem: a low-dimensional manifold $\mathcal{M} \in \mathbb{R}^m$ embedded in high dimensional space $\mathcal{P}\in\mathbb{R}^p$. On this manifold, all the faces lie inside the population hyper-ellipsoid and the embedding of images belonging to each identity or a class are clustered into their own class-specific hyper-ellipsoids. The capacity of this manifold is the number of identities (class-specific hyper-ellipsoids) that can be packed into the population hyper-ellipsoid within an error tolerance or amount of overlap. \label{fig:embedding}}
\end{figure}

The ability to determine this capacity affords the following benefits: (i) Statistical estimates of the upper bound on the number of identities the face representation can resolve. This would allow for informed deployment of face recognition systems based on the expected scale of operation; (ii) Estimate the maximal gallery size for the face representation \emph{without} having to exhaustively evaluate the face representation at that scale. Consequently, capacity offers an alternative dataset\footnote{Class of datasets as opposed to a specific dataset.} agnostic metric for comparing different face representations.

An attractive solution for estimating the capacity of face representations is to leverage the notion of packing bounds\footnote{A generalization of the well studied sphere-packing problem.}; the maximal number of shapes that can be fit, without overlapping, within the support of the representation space. A loose bound on this packing problem can be obtained as a ratio of the volume of the support space and the volume of the shape. In the context of face representations, the representation support can be modeled as a low-dimensional population manifold $\mathcal{M}\in\mathbb{R}^m$ embedded within a high-dimensional representation space $\mathcal{P}\in\mathbb{R}^p$, while each class\footnote{In the case of face recognition, each class is an identity (subject) and the number of classes corresponds to the number of identities.} can be modeled as its own manifold $\mathcal{M}_c \subseteq \mathcal{M}$. Under this setting, a bound on the capacity of the representation can be obtained as a ratio of the volumes of the population and class-specific manifolds. However, adopting this approach to obtain empirical estimates of the capacity presents the following challenges:

\begin{enumerate}
\item Estimating the support of the population manifold $\mathcal{M}$ and the class-specific manifolds $\mathcal{M}_c$, especially for a high-dimensional embedding, such as a face representation (typically, several hundred), is an open problem.

\item Estimating the density of the manifolds while accounting for the different sources of noise is a challenging task. In the context of face representations, all the components of a typical face representation pipeline (see Fig. \ref{fig:pipeline}) are potential sources of noise.

\item Obtaining reliable estimates of the volume of arbitrarily shaped high-dimensional manifolds (for capacity bound), is another open problem.
\end{enumerate}

In this paper, we propose a framework that addresses the aforementioned challenges to obtain reliable estimates of the capacity of any face representation. Our solution relies on; (1) modeling the face representation as a low-dimensional Euclidean manifold embedded within a high-dimensional space, (2) projecting and unfolding the manifold to a low-dimensional space, (3) approximating the population manifold by a multivariate Gaussian distribution (equivalently, hyper-ellipsoidal support) in the unfolded low-dimensional space, (4) approximating the class-specific manifolds by a multi-variate Gaussian distribution and estimating its support as a function of the specified FAR, and (5) estimating the capacity as a ratio of the volumes of the population and class-specific hyper-ellipsoids. Figure \ref{fig:embedding} provides a pictorial illustration of the geometrical structure of our setting.

At the core, we leverage advances in deep neural networks (DNNs) for multiple aspects of our solution, relying on their ability to approximate complex non-linear mappings. Firstly, we utilize DNNs to approximate the non-linear function for projecting and unfolding the high-dimensional face representation into a low-dimensional representation, while preserving the local geometric structure of the manifold. Secondly, we utilize DNNs to aid in approximating the density and support of the low-dimensional manifold in the form of multivariate Gaussian distributions as a function of the desired FAR. The key technical contributions of this paper are:

\begin{enumerate}
\item Explicitly accounting for and modeling the manifold structure of the face representation in the capacity estimation process. This is achieved through a DNN based non-linear projection and unfolding of the representation into an equivalent low-dimensional Euclidean space while preserving the local geometric structure of the manifold.

\item A noise model for facial embeddings that explicitly accounts for two sources of uncertainty, uncertainty due to data and the uncertainty in the parameters of the representation function.

\item Establishing a relationship between the support of the class-specific manifolds and the discriminant function of a nearest neighbor classifier. Consequently, we can estimate capacity as a function of the desired operating point, in terms of the maximum desired probability of false acceptance error.

\item The first practical attempt at estimating the capacity of DNN based face representations. We consider two such representations, namely, FaceNet \cite{schroff2015facenet} and SphereFace \cite{liu2017sphereface} consisting of 128-dimensional and 512-dimensional feature vectors, respectively.
\end{enumerate}

Numerical experiments suggest that our proposed model can provide reasonable estimates of capacity. For FaceNet and SphereFace, the upper bounds on the capacity are $3.5\times10^5$ and $1.1\times10^4$, respectively, for LFW \cite{huang2007labeled}, and $2.2\times10^3$ and $3.6\times10^3$, respectively, for IJB-C \cite{maze2018iarpa} at a FAR of 0.1\%. This implies that, on average, the representation should have a true accept rate (TAR) of 100\% at FAR of 0.1\% for $2.2\times10^3$ and $3.6\times10^5$ subject identities for the more challenging IJB-C database and relative less challenging LFW database (see Figures \ref{fig:datasets-lfw} and \ref{fig:datasets-ijbc} for examples of faces in LFW and IJB-C). As such, the capacity estimates represent an upper bound on the maximal scalability of a given face representation. However, empirically, the FaceNet representation only achieves a TAR of 43\% at a FAR of 0.1\% on the IJB-C dataset with 3,531 subjects and a TAR of 94\% at a FAR of 0.1\% on the LFW dataset with 5,749 subjects.


\section{Related Work}

The focus of a majority of the face recognition literature has been on the accuracy of facial recognition on benchmark datasets. In contrast, our goal in this paper is to characterize the maximal discriminative capacity of a given face representation at a specified error tolerance.

A number of approaches have been proposed to analyze various performance metrics of biometric recognition systems, primarily using information theoretic concepts. Schmid \etal \cite{schmid2004performance, schmid2006performance} derive analytical bounds on the probability of error and capacity of biometric systems through large deviation analysis on the distribution of the similarity scores. Bhatnagar \etal \cite{bhatnagar2009estimating} formulated performance indices for biometric authentication. They obtained the capacity of a biometric system following Shannon's channel capacity formulation along with a rate-distortion theory framework to estimate the FAR. Similarly, Wang \etal \cite{wang2007modeling} proposed an approach to model and predict the performance of a face recognition system based on an analysis of the similarity scores. The common theme across this entire body of work is that the performance bounds of these systems are analyzed purely based on the similarity scores obtained as part of the matching process. In contrast, our work in this paper directly analyzes the geometry of the representation space of face recognition systems.

In the context of estimating low-dimensional approximations of data manifolds, many approaches have been proposed. These include Principal Component Analysis \cite{jolliffe1986principal}, Multidimensional Scaling (MDS) \cite{kruskal1964multidimensional}, Laplacian Eigenmaps \cite{belkin2003laplacian}, Locally Linear Embedding \cite{roweis2000nonlinear}, Isomap \cite{tenenbaum2000global} and deep neural network based approaches such as deep autoencoders \cite{hinton2006reducing} and denoising autoencoders \cite{vincent2010stacked}. The main drawback of these approaches is that none of them allude to the dimensionality of the representation manifold, and instead requires it to be manually specified or is typically estimated through ad-hoc approaches. Furthermore, these approaches may not be able to fully preserve the local geometric structure of the manifold at high levels of compression. Recently, Gong \etal \cite{gong2018deepmds} partly addressed both of these challenges in the context of deep neural network based image representations. They estimated the intrinsic dimensionality of the representation and also proposed a method to learn a non-linear mapping that preserves the discriminative performance of the representation to a large extent. We base our manifold unfolding solution on this approach, but with the goal of estimating the capacity as opposed to preserving the discriminative performance of the representation. Therefore, while the latter does not necessitate preserving the local geometric structure of the manifold, the former is critically dependent on the ability of the dimensionality reduction technique to preserve the local geometric structure of the manifold.

In the context of estimating distributions, Gaussian Processes \cite{rasmussen2006gaussian} are a popular and powerful tool to model distributions over functions, offering nice properties such as uncertainty estimates over function values, robustness to over-fitting, and principled ways for hyper-parameter tuning. A number of approaches have been proposed for modeling uncertainties in deep neural networks \cite{gal2015dropout,gal2016theoretically,gal2015bayesian}. Along similar lines, Kendall \etal \cite{kendall2017uncertainties} studied the benefits of explicitly modeling \emph{epistemic}\footnote{Uncertainty due to lack of information about a process.} (model) and \emph{aleatoric} \footnote{Uncertainty stemming from the inherent randomness of a process.} (data) uncertainties \cite{der2009aleatory} in Bayesian deep neural networks for semantic segmentation and depth estimation tasks. Drawing inspiration from this work, we account for these two sources of uncertainties in the process of mapping a normalized facial image into a low-dimensional face representation.

Capacity estimates to determine the uniqueness of other biometric modalities, namely fingerprints and iris have been reported. Pankanti \etal \cite{pankanti2002individuality} derived an expression for estimating the probability of a false correspondence between minutiae-based representations from two arbitrary fingerprints belonging to two different fingers. Zhu \etal \cite{zhu2007statistical} later developed a more realistic model of fingerprint individuality through a finite mixture model to represent the distribution of minutiae in fingerprint images, including minutiae clustering tendencies and dependencies in different regions of the fingerprint image domain. Daugman \cite{daugman2016information} proposed an information theoretic approach to compute the capacity of IrisCode. He first developed a generative model of IrisCode based on Hidden Markov Models and then estimated the capacity of IrisCode by calculating the entropy of this generative model. Adler \etal \cite{adler2009towards} proposed an information theoretic approach to estimate the average information contained within a face representation like Eigenfaces \cite{turk1991face}.

To the best of our knowledge, no such capacity estimation models have been proposed in the literature for representations of faces. Moreover, the distinct nature of representations for fingerprint\footnote{An unordered collection of minutiae points.}, iris\footnote{A binary representation, called the iris code.} and face\footnote{A fixed-length vector of real values.} traits does not allow capacity estimation approaches to carry over from one biometric modality to another. Therefore, we believe that a new model is necessary to establish the capacity of face representations.


\section{Capacity of Face Representations}
\begin{figure*}[!ht]
    \centering
    \includegraphics[width=0.95\textwidth]{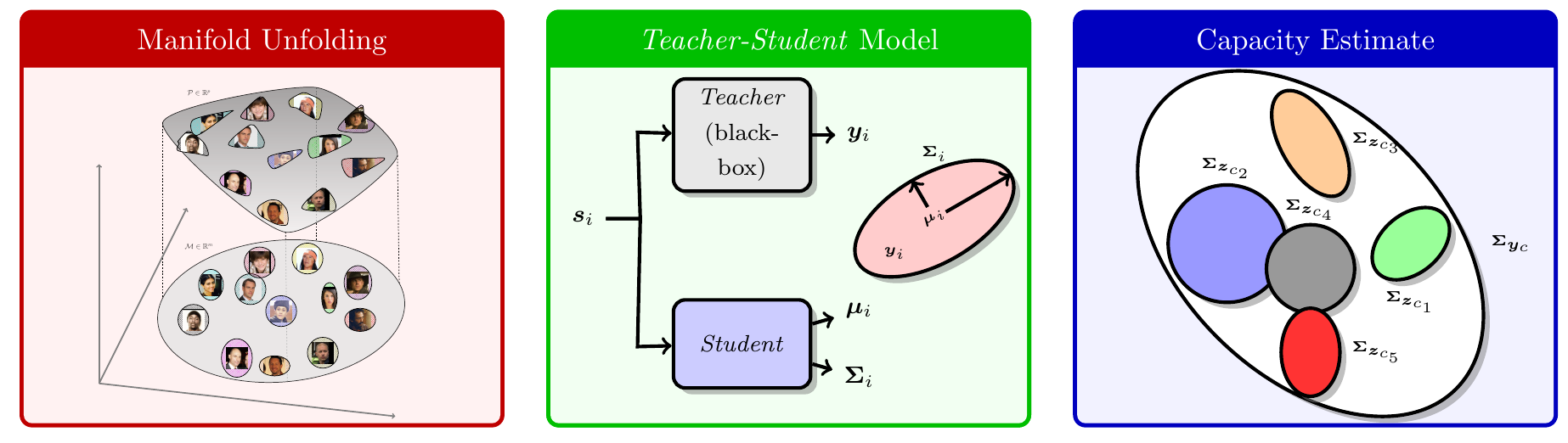}
    \caption{\textbf{Overview of Face Representation Capacity Estimation:} We cast the capacity estimation process in the framework of the sphere packing problem on a low-dimensional manifold. To generalize the sphere packing problem, we replace spheres by hyper-ellipsoids, one per class (subject). Our approach involves three steps; (i) Unfolding and mapping the manifold embedded in high-dimensional space onto a low-dimensional space. (ii) \emph{Teacher}-\emph{Student} model to obtain explicit estimates of the uncertainty (noise) in the embedding due to data as well as the parameters of the representation, and (iii) The uncertainty estimates are leveraged to approximate the density manifold via multi-variate normal distributions (to keep the problem and its analysis tractable), which in turn facilitates an empirical estimate of the capacity of the \emph{teacher} face representation as a ratio of hyper-ellipsoidal volumes.}
    \label{fig:overview}
\end{figure*}

We first describe the setting of the problem and then describe our solution. A pictorial outline of the approach is shown in Fig. \ref{fig:overview}.

\subsection{Face Representation Model \label{sec:channel}}
A face representation model $M$ is a parametric embedding function that maps a face image $\bm{s}$ of identity $\emph{c}$ to a vector space $\bm{x} \in \mathbb{R}^p$, i.e., $\bm{x}=f_{M}(\bm{s};\bm{\theta}_{\mathcal{P}})$, where $\bm{\theta}_{\mathcal{P}}$ is the set of parameters of the embedding function. For example, in the case of a linear embedding function like Principal Component Analysis (PCA), the parameter set $\bm{\theta}_{\mathcal{P}}$ would represent the eigenvectors. And, in the case of a deep neural network based non-linear embedding function, $\bm{\theta}_{\mathcal{P}}$ represents the parameters of the network.

We model the space occupied by the learned face representation as a low-dimensional population manifold $\mathcal{M}\in\mathbb{R}^m$ embedded within a high-dimensional space $\mathcal{P}\in\mathbb{R}^p$. Under this model the features of a given identity $c$ lie on a manifold $\mathcal{M}_c\subseteq\mathcal{M}$. Directly estimating the support and volumes of these manifolds is a very challenging task, especially since the manifold could be a highly entangled surface in $\mathbb{R}^p$. Therefore, we first learn a mapping that can project and unfold the population manifold onto a low-dimensional space whose density, support and volume can be estimated more reliably. 

We base our solution for projecting and unfolding the manifolding on Multidimensional scaling (MDS) \cite{kruskal1964multidimensional}, a classical mapping technique that attempts to preserve local distances (similarities) between points after embedding them in a low-dimensional space. Given data points $\bm{X}=\{\bm{x}_1,\dots,\bm{x}_n\}$ in the high-dimensional space and  denoting by $\bm{Y}=\{\bm{y}_1,\dots,\bm{y}_n\}$ the corresponding points in the low-dimensional space, the MDS projection is formulated as,
\begin{equation}
    \min \sum_{i < j} \left(d_{H}(\bm{x}_i,\bm{x}_j) - d_{L}(\bm{y}_i,\bm{y}_j)\right)^2
\end{equation}
\noindent where $d_H(\cdot)$ and $d_L(\cdot)$ are distance (similarity) metrics in the high and low dimensional space, respectively. Different choices of the metric, lead to different dimensionality reduction algorithms. For instance, classical metric MDS is based on Euclidean distance between the points while using the geodesic distance induced by a neighborhood graph leads to Isomap \cite{tenenbaum2000global}. Similarly, many different distance metrics have been proposed corresponding to non-linear mappings between the ambient space and the intrinsic space. A majority of these approaches are based on spectral decompositions and suffer a number of drawbacks: (i) computational complexity scales as $\mathcal{O}(n^3)$ for $n$ data points, (ii) ambiguity in the choice of the correct non-linear function, and (iii) lack of an explicit mapping function for unseen data samples due to their iterative nature.

\begin{figure*}[t]
    \centering
    \includegraphics[width=\textwidth]{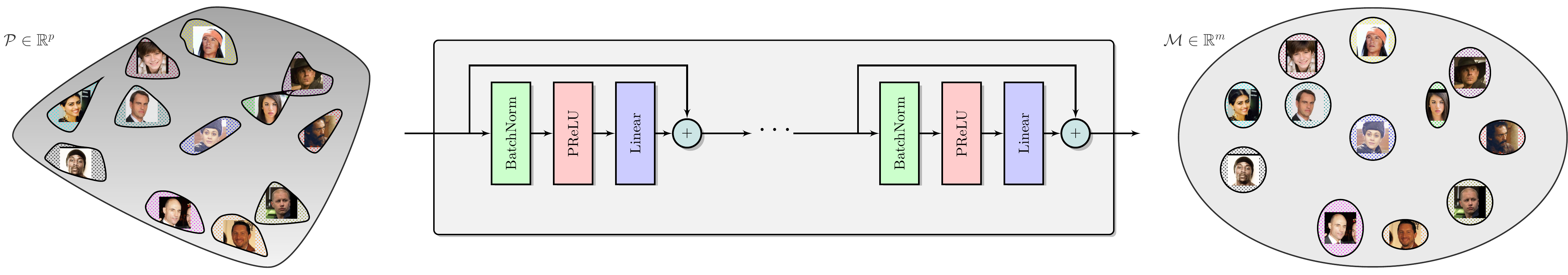}
    \caption{\textbf{Manifold Unfolding:} A DNN based non-linear mapping is learned to unfold and project the population manifold into a lower dimensional space. The network is optimized to preserve the geodesic distances between pairs of points in the high and low dimensional space.}
    \label{fig:nlle}
\end{figure*}

To overcome these limitations, we employ a DNN to approximate the non-linear mapping that transforms the population manifold in high-dimensional space, $\bm{x} \in \mathbb{R}^p$, to the unfolded manifold in low-dimensional space, $\bm{y} \in \mathbb{R}^m$ by a parametric function $\bm{y}=f_{P}(\bm{x}; \bm{\theta}_{\mathcal{M}})$ with parameters $\bm{\theta}_{\mathcal{M}}$. We learn the parameters of the mapping within the MDS framework to minimize the following objective,
\begin{equation}
    \footnotesize{
    \begin{aligned}
    \min_{\bm{\theta}_{\mathcal{M}}} \sum_{i < j} \left[d_H(\bm{x}_i,\bm{x}_j) - d_L(f(\bm{x}_i;\bm{\theta}_{\mathcal{M}}),f(\bm{x}_j;\bm{\theta}_{\mathcal{M}}))\right]^2 + \lambda\|\bm{\theta}_{\mathcal{M}}\|_2^2
    \end{aligned}}
    \label{eq:deepmds}
\end{equation}
\noindent where the second term is a regularizer with a hyperparameter $\lambda$. Since our primary goal is to estimate the capacity of the representation we map the manifold into the low-dimensional space while preserving the local geometry of the manifold in the form of pairwise distances. To achieve this goal, we choose $d_H(\bm{x}_i,\bm{x}_j)=1+\frac{\bm{x}_i^T\bm{x}_j}{\|\bm{x}_i\|_2\|\bm{x}_j\|_2}$ corresponding to the cosine distance between the features in the high dimensional space and $d_L(\bm{y}_i,\bm{y}_j)=\|\bm{y}_i-\bm{y}_j\|_2$ corresponding to the euclidean distance in the low dimensional space. Figure \ref{fig:nlle} shows an illustration of the DNN based mapping.

\subsection{Estimating Uncertainties in Representations \label{sec:uncertainty}}
The projection model learned in the previous section can be used to obtain the population manifold by propagating multiple images from many identities through it. However, this process only provides point estimates (samples) from the manifold and does not account for the uncertainty in the manifold. Accurately estimating the capacity of the face representation, however, necessitates modeling the uncertainty in the representation stemming from different sources of noise in the process of extracting feature representations from a given facial image.

A probabilistic model for the space of noisy embeddings $\bm{y}$ generated by a black-box facial representation model (\emph{teacher}\footnote{We adopt the terminology of teacher-student models from the model compression community\cite{ba2014deep}.}) $M_t$ with parameters $\bm{\theta} = \{\bm{\theta}_{\mathcal{P}}, \bm{\theta}_{\mathcal{M}}\}$ can be formulated as follows:
\begin{equation}
    \small{
    \begin{aligned}
    p(\bm{y}|\bm{S}^{*},\bm{Y}^{*}) &= \int p(\bm{y}|\bm{s},\bm{S}^*,\bm{Y}^*)p(\bm{s}|\bm{S}^*,\bm{Y}^*)d\bm{s} \\
    &= \int\int p(\bm{y}|\bm{s},\bm{\theta})p(\bm{\theta}|\bm{S}^*,\bm{Y}^*)p(\bm{s}|\bm{S}^*,\bm{Y}^*)d\bm{\theta}d\bm{s}
    \end{aligned}}
\end{equation}
\noindent where $\bm{Y}^*=\{\bm{y}_1,\dots,\bm{y}_N\}$ and $\bm{S}^*=\{\bm{s}_1,\dots,\bm{s}_N\}$ are the training samples to estimate the model parameters $\bm{\theta}$, $p(\bm{y}|\bm{s},\bm{\theta})$ is the \emph{aleatoric} (data) uncertainty given a set of parameters, $p(\bm{\theta}|\bm{S}^*,\bm{Y}^*)$ is the \emph{epistemic} (model) uncertainty in the parameters given the training samples and $p(\bm{s}|\bm{S}^*,\bm{Y}^*) \sim \mathcal{N}(\bm{\mu}_g,\bm{\Sigma}_g)$ is the Gaussian approximation (see Section \ref{sec:capacity} for justification) of the underlying manifold of noiseless embeddings. Furthermore, we assume that the true mapping between the image $\bm{s}$ and the noiseless embedding $\bm{\mu}$ is a deterministic but unknown function i.e., $\bm{\mu}=f(\bm{s},\bm{\theta})$.

The black-box nature of the \emph{teacher} model however only provides $\mathcal{D}=\{\bm{s}_i,\bm{y}_i\}_{i=1}^N$, pairs of facial images $\bm{s}_i$ and their corresponding noisy embeddings $\bm{y}_i$, a single sample from the distribution $p(\bm{y}|\bm{S}^*,\bm{Y}^*)$. Therefore, we learn a \emph{student} model $M_s$ with parameters $\bm{w}$ to mimic the \emph{teacher} model. Specifically, the $\emph{student}$ model approximates the data dependent \emph{aleatoric} uncertainty $p(\bm{y}_i|\bm{s}_i,\bm{w}) \sim \mathcal{N}(\bm{\mu}_i,\bm{\Sigma}_i)$, where $\bm{\mu}_i$ represents the data dependent mean of the noiseless embedding and $\bm{\Sigma}_i$ represents the data dependent uncertainty around the mean. This \emph{student} is an approximation of the unknown underlying probabilistic $\emph{teacher}$, by which an input image $\bm{s}$ generates noisy embeddings $\bm{y}$ of ideal noiseless embeddings $\bm{\mu}$, for a given set of parameters $\bm{w}$, i.e., $p(\bm{y}_i|\bm{s}_i,\bm{w}) \approx p(\bm{y}_i|\bm{\mu}_i,\bm{\theta})$. Finally, we employ a variational distribution to approximate the \emph{epistemic} uncertainty of the \emph{teacher} i.e., $p(\bm{w}|\bm{S}^*,\bm{Y}^*) \approx p(\bm{\theta}|\bm{S}^*,\bm{Y}^*)$.

\vspace{4pt}
\noindent \textbf{Learning:} Given pairs of facial images and their corresponding embeddings from the $\emph{teacher}$ model, we learn a $\emph{student}$ model to mimic the outputs of the \emph{teacher} for the same inputs in accordance to the probabilistic model described above. We use parameterized functions, $\bm{\mu}_i = f(\bm{s}_i;\bm{w}_{\bm{\mu}})$ and $\bm{\Sigma}_i = f(\bm{s}_i;\bm{w}_{\bm{\Sigma}})$ to characterize the $\emph{aleatoric}$ uncertainty $p(\bm{y}_i|\bm{s}_i,\bm{w})$, where $\bm{w}=\{\bm{w}_{\mu},\bm{w}_{\Sigma}\}$. We choose deep neural networks, specifically convolutional neural networks as our functions $f(\cdot;\bm{w}_{\bm{\mu}})$ and $f(\cdot;\bm{w}_{\bm{\Sigma}})$. For the \emph{epistemic} uncertainty, while many deep learning based variational inference \cite{kingma2015variational,gal2015dropout,blundell2015weight} approaches have been proposed, we use the simple interpretation of dropout as our variational approximation \cite{gal2015dropout}. Practically, this interpretation simply characterizes the uncertainty in the deep neural network weights $\bm{w}$ through a Bernoulli sampling of the weights.

We learn the parameters of our probabilistic model $\bm{\phi}=\{\bm{w}_{\bm{\mu}},\bm{w}_{\bm{\Sigma}},\bm{\mu}_g,\bm{\Sigma}_{g}\}$ through maximum-likelihood estimation i.e., minimizing the negative log-likelihood of the observations $\bm{Y}=\{\bm{y}_1,\dots,\bm{y}_N\}$. This translates to optimizing a combination of loss functions:
\begin{eqnarray}
    \min_{\bm{\phi}} && \mathcal{L}_{s} + \lambda\mathcal{L}_{g} + \gamma\mathcal{L}_{r_s} + \delta\mathcal{L}_{r_g}
    \label{eq:all}
\end{eqnarray}
where $\lambda$, $\gamma$ and $\delta$ are the weights for the different loss functions, $\mathcal{L}_{r_s} = \frac{1}{2N}\sum_{i=1}^N\|\bm{\Sigma}_i\|_F^2$ and $\mathcal{L}_{r_g} = \frac{1}{2}\|\bm{\Sigma}_g\|_F^2$ are the regularization terms and $\mathcal{L}_s$ is the loss function of the student that captures the log-likelihood of a given noisy representation $\bm{y}_i$ under the distribution $\mathcal{N}(\bm{\mu}_i,\bm{\Sigma}_i)$.
\begin{eqnarray}
    \mathcal{L}_{s} = \frac{1}{2} \sum_{i=1}^N\ln|\bm{\Sigma}_i| + \frac{1}{2} \text{Trace}\left(\sum_{i=1}^N\bm{\Sigma}_i^{-1}\left[(\bm{y}_i-\bm{\mu}_i)(\bm{y}_i-\bm{\mu}_i)^T\right]\right) \nonumber
    \label{eq:student}
\end{eqnarray}
\noindent $\mathcal{L}_g$ is the log-likelihood of the population manifold of the embedding under the approximation by a multi-variate normal distribution $\mathcal{N}(\bm{\mu}_g,\bm{\Sigma}_g)$.
\begin{eqnarray}
    \mathcal{L}_{g} = \frac{N}{2} \ln|\bm{\Sigma}_g| + \frac{1}{2} \text{Trace}\left(\bm{\Sigma}_g^{-1}\sum_{i=1}^N\left[(\bm{y}_i-\bm{\mu}_g)(\bm{y}_i-\bm{\mu}_g)^T\right]\right) \nonumber
    \label{eq:global}
\end{eqnarray}
For computational tractability we make a simplifying assumption on the covariance matrix $\bm{\Sigma}$ by parameterizing it as a diagonal matrix i.e., the off-diagonal elements are set to zero. This parametrization corresponds to independence assumptions on the uncertainty along each dimension of the embedding. The sparse parametrization of the covariance matrix yields two computational benefits in the learning process. Firstly, it is sufficient for the \emph{student} to predict only the diagonal elements of the covariance matrix. Secondly, positive semi-definitiveness constraints on a diagonal matrix can be enforced simply by forcing all the diagonal elements of the matrix to be non-negative. To enforce non-negativity on each of the diagonal variance values, we predict the log variance, $l_j=\log \sigma_j^2$. This allows us to re-parameterize the \emph{student} likelihood in terms of $\bm{l}_i$:
\begin{dmath}	
    \mathcal{L}_{s} = \frac{1}{2} \sum_{i=1}^N\sum_{j=1}^d l_i^j + \frac{1}{2}\sum_{i=1}^N\sum_{j=1}^d\frac{\left(y_i^j-\mu_i^j\right)^2}{\exp\left(l_i^j\right)}
    \label{eq:student_psd_diag}
\end{dmath}
\noindent Similarly, we reparameterize the likelihood of the noiseless embedding as a function of $\bm{l}_g$, the log variance along each dimension. The regularization terms are also reparameterized as, $\mathcal{L}_{r_s} = \frac{1}{2N}\sum_{i=1}^N\sum_{j=1}^d \exp\left(l_i^j\right)$ and $\mathcal{L}_{r_g} = \frac{1}{2}\sum_{j=1}^d \exp\left(l_g^j\right)$. We empirically estimate $\bm{\mu}_g$ as $\bm{\mu}_g=\frac{1}{N}\sum_{i=1}^N\bm{y}_i$ and the other parameters $\bm{\phi}=\{\bm{w}_{\bm{\mu}},\bm{w}_{\bm{\Sigma}},\bm{\Sigma}_{g}\}$ through stochastic gradient descent \cite{bottou2010large}. The gradients of the parameters are computed by backpropagating \cite{rumelhart1988learning} the gradients of the outputs through the network.

\vspace{4pt}
\noindent \textbf{Inference:}
The \emph{student} model that has been learned can now be used to infer the uncertainty in the embeddings of the original \emph{teacher} model. For a given facial image $\bm{s}$, the \emph{aleatoric} uncertainty can be predicted by a feed-forward pass of the image $\bm{s}$ through the network i.e., $\bm{\mu}=f(\bm{s},\bm{w}_{\bm{\mu}})$ and $\bm{\Sigma}=f(\bm{s},\bm{w}_{\bm{\Sigma}})$. The \emph{epistemic} uncertainty can be approximately estimated through Monte-Carlo integration over different samples of model parameters $\bm{w}$. In practice the parameter sampling is performed through the use of dropout at inference. In summary, the total uncertainty in the embedding of each facial image $\bm{s}$ is estimated by performing Monte-Carlo integration over a total of $T$ evaluations,
\begin{eqnarray}
    \bm{\hat{\mu}}_i &=& \frac{1}{T}\sum_{t=1}^T\bm{\mu}_i^t\\
    \bm{\hat{\Sigma}}_i &=& \frac{1}{T}\sum_{t=1}^T\left(\bm{\mu}_i^t-\bm{\hat{\mu}}_i\right)\left(\bm{\mu}_i^t-\bm{\hat{\mu}}_i\right)^T + \frac{1}{T}\sum_{t=1}^T\bm{\Sigma}_i^t 
\end{eqnarray}
\noindent where $\bm{\mu}_i^t$ and $\bm{\Sigma}_i^t$ are the predicted \emph{aleatoric} uncertainty for each feed-forward evaluation of the network.

\subsection{Manifold Approximation \label{sec:capacity}}
The student model described in Section \ref{sec:uncertainty} allows us to extract uncertainty estimates of each individual image. Given these estimates the next step is to estimate the density and support of the population and class-specific low-dimensional manifolds. 

Multiple existing techniques can be employed for this purpose under different modeling assumptions, ranging from non-parametric models like kernel density estimators and convex-hulls to parametric models like multivariate Gaussian distribution and escribed hyper-spheres. The non-parametric and parametric models span the trade-off between the accuracy of the manifold's shape estimate and the computational complexity of fitting the shape and calculating the volume of the manifold. While the non-parametric models provide more accurate estimates of the density and support of the manifold, the parametric models potentially provide more robust and computationally tractable estimates of the density and volume of the manifolds. For instance, estimating the convex hull of samples in high-dimensional space and its volume is both computationally prohibitive and less robust to outliers.

To overcome the aforestated challenges we approximate the density of the population and class-specific manifolds in the low-dimensional space via multi-variate normal distributions. The choice of the normal distribution approximation is motivated by multiple factors; (a) probabilistically it leads to a robust and computationally efficient estimate of the density of the manifold, (b) geometrically it leads to a hyper-ellipsoidal approximation of the manifold, which in turn allows for efficient and exact estimates of the support and volume of the manifold as a function of the desired false acceptance rate (see Section \ref{sec:decision_theory}), and (c) the low-dimensional manifold obtained through projection and unfolding of the high-dimensional representation is implicitly designed, through Eq. \ref{eq:deepmds}, to cluster the facial images belonging to the same identity, and therefore a normal distribution is a realistic (see Section \ref{sec:toy-example}) approximation of the manifold.

Empirically we estimate the parameters of these distributions as follows. The mean of the population embedding is computed as $\bm{\mu}_{\bm{y}_c} = \frac{1}{C}\sum_{c=1}^C \bm{\hat{\mu}}^{c}$, where $\bm{\hat{\mu}}^c = \frac{1}{N_c}\sum_{i=1}^{N_c}\bm{\hat{\mu}}^c_i$. The covariance of the population embedding $\bm{\Sigma}_{\bm{y}_c}$ is estimated as,
\begin{eqnarray}
  \bm{\tilde{\Sigma}}^c=\argmax _{\bm{\hat{\Sigma}}^c}\left|\bm{\hat{\Sigma}}^c + \frac{1}{C}\sum_{c=1}^C(\bm{\hat{\mu}}^c-\bm{\mu}_{\bm{y}_c})(\bm{\hat{\mu}}^c-\bm{\mu}_{\bm{y}_c})^T\right| \nonumber \\
    \bm{\Sigma}_{\bm{y}_c}+\bm{\Sigma}_{\bm{z}_c}=\bm{\tilde{\Sigma}}^c+\frac{1}{C}\sum_{c=1}^C(\bm{\hat{\mu}}^c-\bm{\mu}_{\bm{y}_c})(\bm{\hat{\mu}}^c-\bm{\mu}_{\bm{y}_c})^T
    \label{eq:population}
\end{eqnarray}
\noindent where $\bm{\hat{\Sigma}}^c=\frac{1}{N_c}\sum_{i=1}^{N_c}\bm{\hat{\Sigma}}_i^c$. Along the same lines, the class-specific covariance $\bm{\Sigma}_{\bm{z}_c}$ of a class $c$ is estimated as,
\begin{eqnarray}
    \bm{\Sigma}_{\bm{z}_c} &=& \frac{1}{N_cT}\sum_{i=1}^{N_c}\sum_{t=1}^T\left[\left(\bm{\mu}_i^t-\bm{\hat{\mu}}_i\right)\left(\bm{\mu}_i^t-\bm{\hat{\mu}}_i\right)^T+ \bm{\Sigma}_i^t \right]
    \label{eq:class-specific}
\end{eqnarray}

\subsection{Decision Theory and Model Capacity \label{sec:decision_theory}}
\begin{figure*}[t]
    \centering
    \includegraphics[width=0.95\textwidth]{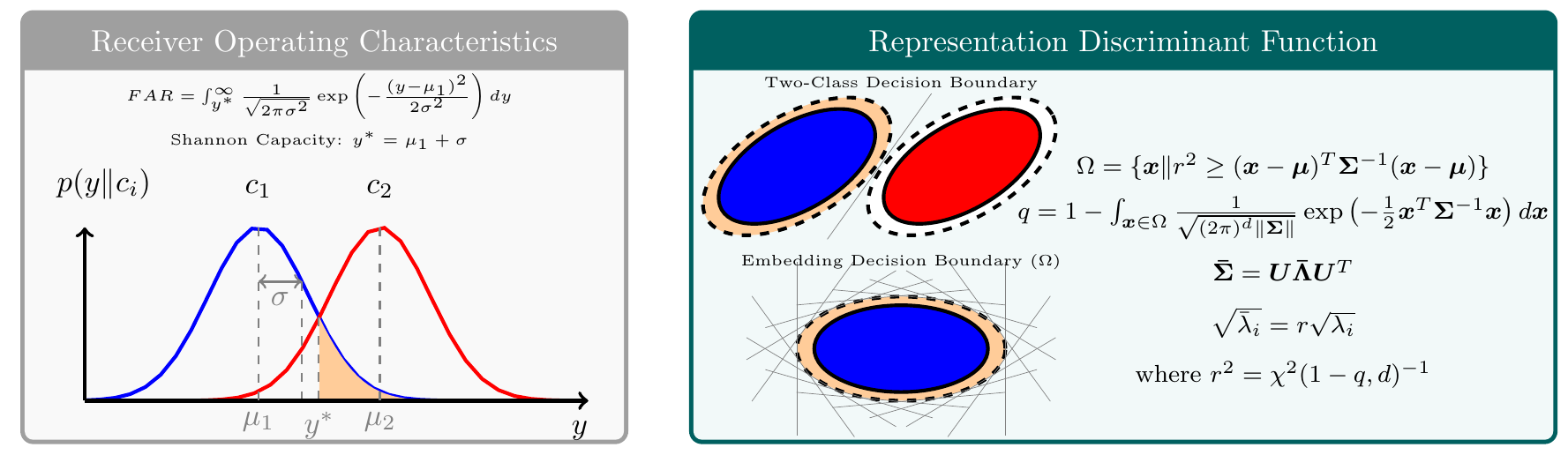}
    \caption{\textbf{Decision Theory and Capacity:} We illustrate the relation between capacity and the discriminant function corresponding to a nearest neighbor classifier. \textbf{Left:} Depiction of the notion of decision boundary and probability of false accept between two identical one dimensional Gaussian distributions. Shannon's definition of capacity corresponds to the decision boundary being one standard deviation away from the mean. \textbf{Right:} Depiction of the decision boundary induced by the discriminant function of nearest neighbor classifier. Unlike in the definition of Shannon's capacity, the size of the ellipsoidal decision boundary is determined by the maximum acceptable false accept rate. The probability of false acceptance can be computed through the cumulative distribution function of a $\chi^2(r^2,d)$ distribution.}
    \label{fig:decision_theory}
\end{figure*}

Thus far, we developed the tools necessary to characterize the face representation manifold and estimate its density. In this section we will determine the support and volume of the population and class-specific manifolds as a function of the specified false accept rate (FAR).

Our representation space is composed of two components: the population manifold of all the classes approximated by a multi-variate Gaussian distribution and the embedding noise of each class approximated by a multi-variate Gaussian distribution. Under these settings, the decision boundaries between the classes that minimizes the classification error rate are determined by discriminant functions \cite{duda2012pattern}. As illustrated in Fig. \ref{fig:decision_theory}, for a two-class problem, the discriminant function is a hyper-plane in $\mathbb{R}^d$, with the optimal hyper-plane being equidistant from both the classes. Moreover, the separation between the classes determines the operating point and hence the FAR. In the multi-class setting the optimal discriminant function is the surface encompassed by all the pairwise hyper-planes, which asymptotically reduces to a high-dimensional hyper-ellipsoid. The support of this enclosing hyper-ellipsoid can be determined by the desired operating point in terms of the maximal error probability of false acceptance.

Under the multi-class setting, the capacity estimation problem is equivalent to the geometrical problem of ellipse packing, which seeks to estimate the maximum number of small hyper-ellipsoids that can be packed into a larger hyper-ellipsoid. In the context of face representations the small hyper-ellipsoids correspond to the class-specific enclosing hyper-ellipsoids as described above while the large hyper-ellipsoid corresponds to the space spanned by the population of all classes. The volume $V$ of a hyper-ellipsoid corresponding to a Mahalanobis distance $r^2=(\bm{x}-\bm{\mu})^T\bm{\Sigma}^{-1}(\bm{x}-\bm{\mu})$ with covariance matrix $\bm{\Sigma}$ is given by the following expression, $V=V_d|\bm{\Sigma}|^{\frac{1}{2}}r^d$, where $V_d$ is the volume of the $d$-dimensional hypersphere. An upper bound on the capacity of the face representation can be computed simply as the ratio of the volumes of the population and the class-specific hyper-ellipsoids,
\begin{equation}
    \label{eq:volume}
    \begin{aligned}
    C &\leq \left(\frac{V_{\bm{y}_c,\bm{z}_c}}{V_{\bm{z}_c}}\right) \\
    &= \left(\frac{V_d|\bm{\Sigma}_{\bm{y}_c}+\bm{\Sigma}_{\bm{z}_c}|^{\frac{1}{2}}r_{\bm{y}_c}^d}{V_d|\bm{\Sigma}_{\bm{z}_c}|^{\frac{1}{2}}r_{\bm{z}_c}^d}\right) = \left(\frac{|\bm{\Sigma}_{\bm{y}_c}+\bm{\Sigma}_{\bm{z}_c}|^{\frac{1}{2}}r_{\bm{y}_c}^d}{|\bm{\Sigma}_{\bm{z}_c}|^{\frac{1}{2}}r_{\bm{z}_c}^d}\right) \\
    &= \left(\frac{|\bm{\bar{\Sigma}}_{\bm{y}_c,\bm{z}_c}|^{\frac{1}{2}}}{|\bm{\bar{\Sigma}}_{\bm{z}_c}|^{\frac{1}{2}}}\right)
    \end{aligned}
\end{equation}
\noindent where $V_{\bm{y}_c,\bm{z}_c}$ is the volume of population hyper-ellipsoid and $V_{\bm{z}_c}$ is the volume of the class-specific hyper-ellipsoid. The size of the population hyper-ellipsoid $r_{\bm{y}_c}$ is chosen such that a desired fraction of all the classes lie within the hyper-ellipsoid and $r_{\bm{z}_c}$ determines the size of the class-specific hyper-ellipsoid. $\bm{\bar{\Sigma}}_{\bm{y}_c,\bm{z}_c}$ and $\bm{\bar{\Sigma}}_{\bm{z}_c}$ are the effective sizes of the enclosing population and class-specific hyper-ellipsoids respectively. For each of the hyper-ellipsoids the effective radius along the $i$-th principal direction is $\sqrt{\bar{\lambda}_i} = r\sqrt{\lambda_i}$, where $\sqrt{\lambda}_i$ is the radius of the original hyper-ellipsoid along the same principal direction.
 
This geometrical interpretation of the capacity reduces to the Shannon capacity \cite{cover2012elements} when $r_{\bm{y}_c}$ and $r_{\bm{z}_c}$ are chosen to be the same i.e., when $r_{\bm{y}_c}=r_{\bm{z}_c}$. Consequently, in this instance, the choice of $r_{\bm{y}_c}$ for the population hyper-ellipsoid implicitly determines the boundary of separation between the classes and hence the operating false accept rate (FAR) of the embedding. For instance, when computing the Shannon capacity of the face representation choosing $r_{\bm{y}_c}$ such that 95\% of the classes are enclosed within the population hyper-ellipsoid would implicitly correspond to operating at a FAR of 5\%.  However, practical face recognition systems need to operate at lower false accept rates, dictated by the desired level of security.

The geometrical interpretation of the capacity described in Eq. \ref{eq:volume} directly enables us to compute the representation capacity as a function of the desired operating point as determined by its corresponding false accept rate. The size of the population hyper-ellipsoid $r_{\bm{y}_c}$ will be determined by the desired fraction of classes to enclose or alternatively other geometric shapes like the minimum volume enclosing hyper-ellipsoid or the maximum volume inscribed hyper-ellipsoid of a finite set of classes, both of which correspond to a particular fraction of the population distribution. Similarly, the desired false accept rate $q$ determines the size of the class-specific hyper-ellipsoid $r_{\bm{z}_c}$.

Let $\Omega=\{\bm{x} \mid r^2 \geq (\bm{x}-\bm{\mu})^T\bm{\Sigma}^{-1}(\bm{x}-\bm{\mu})\}$ be the enclosing hyper-ellipsoid. Without loss of generality, assuming that the class-specific hyper-ellipsoid is centered at the origin, the false accept rate $q$ can be computed as,
\begin{eqnarray}
    q &=& 1-\int_{\bm{x} \in \Omega} \frac{1}{\sqrt{(2\pi)^{d}|\bm{\Sigma}|}}\exp\left(-\frac{\bm{x}^T\bm{\Sigma}^{-1}\bm{x}}{2}\right)d\bm{x}
\end{eqnarray}

\noindent Reparameterizing the integral as $\bm{y} = \bm{\Sigma}^{-\frac{1}{2}}\bm{x}$, we have $\Omega=\{\bm{y} \mid r^2\geq \bm{y}^T\bm{y}\}$ and,
\begin{equation}
    q = 1-\int_{\bm{y}\in\Omega} \frac{1}{\sqrt{(2\pi)^{d}}}\exp\left(-\frac{\bm{y}^T\bm{y}}{2}\right)d\bm{y}
    \label{eq:chi-squared}
\end{equation}
\noindent where $\{y_1, \dots, y_n\}$ are independent standard normal random variables. The Mahalanobis distance $r^2$ is distributed according to the $\chi^2(r^2,d)$ distribution with $d$ degrees of freedom and $1-q$ is the cumulative distribution function of $\chi^2(r^2,d)$. Therefore, given the desired FAR $q$, the corresponding Mahalanobis distance $r_{\bm{z}_c}$ can be obtained from the inverse CDF of the $\chi^2(r^2_{\bm{z}},d)$ distribution. Along the same lines, the size of the population hyper-ellipsoid $r_{\bm{y}_c}$ can be estimated from the inverse CDF of the $\chi^2(r^2_{\bm{y}},d)$ distribution given the desired fraction of classes to encompass. These estimates of $r_{\bm{z}_c}$ and $r_{\bm{y}_c}$ can be utilized in Eq. \ref{eq:volume} to estimate the capacity as a function of the desired FAR. Algorithm \ref{algo:capacity} provides a high-level outline of our complete capacity estimation procedure.

\begin{algorithm}
\protect\caption{\label{algo:capacity} Face Representation Capacity Estimation}
\begin{algorithmic}
\STATE \textbf{Input:} Representation $f_{M}(\cdot, \bm{\theta}_{\mathcal{P}})$, a face dataset and desired FAR.
\STATE \textbf{Output:} Capacity estimate at specified FAR.
\STATE \textbf{Step 1:} Learn parametric mapping $f_{P}(\cdot,\bm{\theta}_{\mathcal{M}}): \bm{x}\rightarrow\bm{y}$ (Eq. \ref{eq:deepmds})
\STATE \textbf{Step 2:} Learn \emph{student} model $M_s$ to mimic and provide uncertainty estimates of \emph{teacher} $M_t=(M,P)$ (Eq. \ref{eq:all})
\STATE \textbf{Step 3:} Estimate density and support $\bm{\Sigma}_{\bm{y}_c}$ of population manifold (Eq. \ref{eq:population})
\STATE \textbf{Step 4:} Estimate density and support $\bm{\Sigma}_{\bm{z}_c}$ of class-specific manifolds (Eq. \ref{eq:class-specific})
\STATE \textbf{Step 5:} Obtain $r_{\bm{y}_c}$ and $r_{\bm{z}_c}$ for desired population fraction and FAR, respectively (Eq. \ref{eq:chi-squared})
\STATE \textbf{Step 6:} Obtain capacity estimate for desired population fraction and FAR using $r_{\bm{y}_c}$ and $r_{\bm{z}_c}$ (Eq. \ref{eq:volume})
\end{algorithmic}
\end{algorithm}


\section{Numerical Experiments \label{sec:experiments}}
In this section we will, (a) illustrate the capacity estimation process on a two-dimensional toy example, (b) estimate the capacity of a deep neural network based face representation model, specifically FaceNet and SphereFace on multiple datasets of increasing complexity, and (c) study the effect of different design choices of the proposed capacity estimation approach.

\subsection{Two-Dimensional Toy-Example \label{sec:toy-example}}

\begin{figure}[h]
    \centering
    \includegraphics[width=0.5\textwidth]{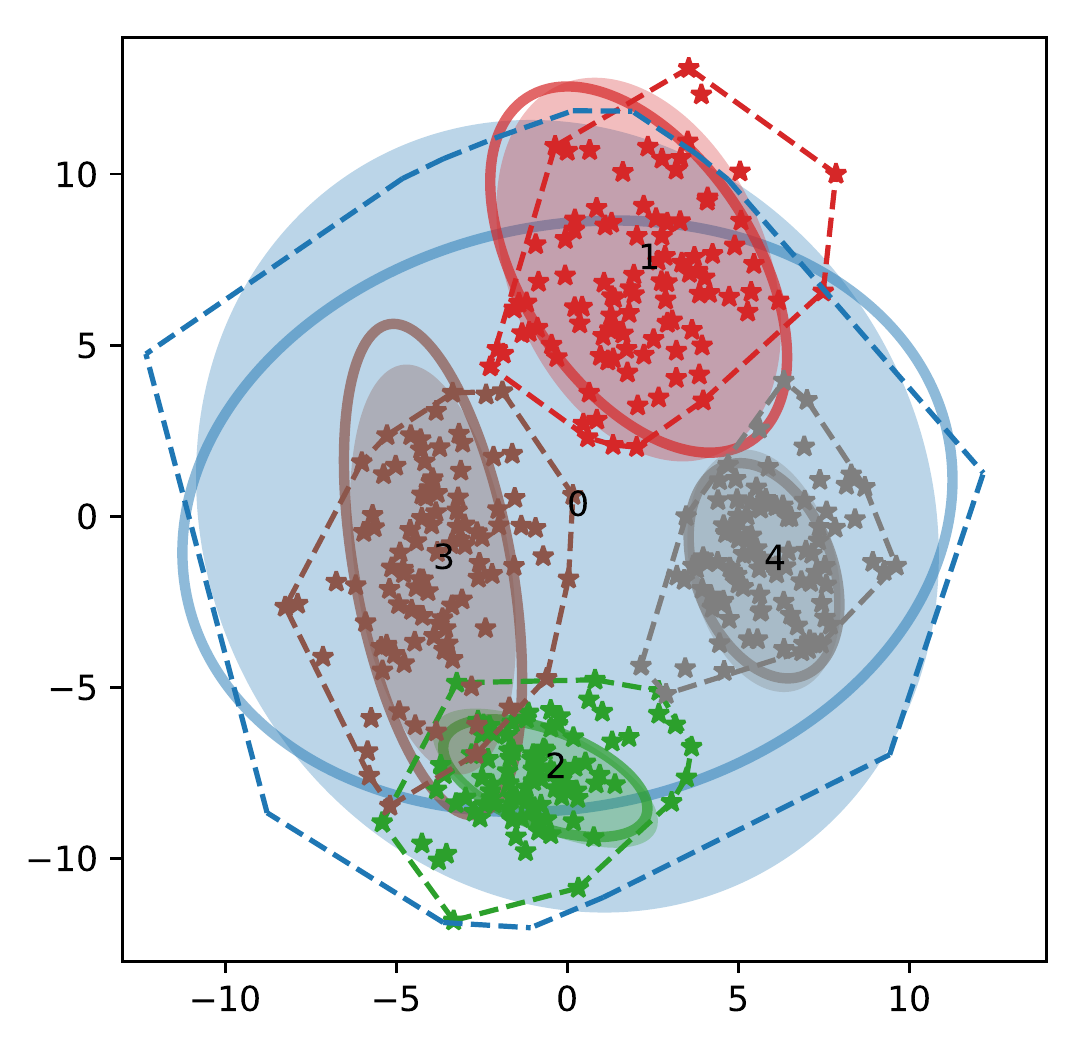}
    \caption{\textbf{Sample Representation Space:} Illustration of a two-dimensional space where the underlying population and class-specific representations (we show four classes) are 2-D Gaussian distributions (solid ellipsoids). Samples from the classes (colored $\star$) are utilized to obtain estimates of this underlying population and class-specific distributions (solid lines). As a comparison, the support of the samples in the form of a convex hull are also shown (dashed lines).}
    \label{fig:toy}
\end{figure}

\begin{table*}[t]
    \centering
    \caption{Capacity of Two-Dimensional Toy Example at 1\% FAR}
    \label{table:toy}
    \scalebox{0.7}{
    \begin{tabular}{cccccccccccccccc}
    \toprule
    \multirow{2}{*}{Manifold} && \multicolumn{5}{c}{Population} && \multicolumn{5}{c}{Class (max area)} & \multirow{2}{*}{Estimated} & \multirow{2}{*}{Ground-Truth}\\
    \cline{3-7} \cline{9-13} 
    && \multicolumn{2}{c}{Covariance} && \multicolumn{2}{c}{Area} && \multicolumn{2}{c}{Covariance} && \multicolumn{2}{c}{Area} \\
    \cline{3-4} \cline{6-7} \cline{9-10} \cline{12-13} 
    Support && Estimate & Ground-Truth && Estimate & Ground-Truth && Estimate & Ground-Truth && Estimate & Ground-Truth & Capacity & Capacity \\
    
    \midrule
    \multirow{3}{*}{Ellipse} && \multirow{3}{*}{$ \begin{bmatrix} 10.84 & 0.56 \\ 0.56 & 11.57 \end{bmatrix} $} & \multirow{3}{*}{$ \begin{bmatrix} 10.34 & 0.71 \\ 0.71 & 11.79 \end{bmatrix} $} && \multirow{3}{*}{35.15} & \multirow{3}{*}{34.62} && \multirow{3}{*}{$ \begin{bmatrix} 4.96 & 0.47 \\ 0.47 & 6.54 \end{bmatrix} $} & \multirow{3}{*}{$ \begin{bmatrix} 4.18 & 0.97 \\ 0.97 & 5.86 \end{bmatrix} $} && \multirow{3}{*}{17.84} & \multirow{3}{*}{15.25} & \multirow{3}{*}{1.97} & \multirow{3}{*}{2.27} \\
    \\
    \\
    \\
    \multirow{1}{*}{Convex Hull} && \multirow{1}{*}{--} & \multirow{1}{*}{--} && \multirow{1}{*}{403.91} & - &
    & -- & -- && 102.65 & - & 3.93 & 2.27 \\
    \bottomrule
    \end{tabular}}
\end{table*}

We consider an illustrative example to demonstrate the capacity estimation process given a constellation of classes in a two-dimensional representation space. We model the distribution of the population space of classes (class centers to be specific) as a multi-variate normal distribution, while the feature space of each class is modeled as a two-dimensional normal distribution. From this model, we sample 100 different classes from the underlying population distribution and for each of these classes we sample features from the ground truth multi-variate normal distribution for that class. From these samples, we estimate the covariance matrix of the population space distribution and that of the individual classes.

Figure \ref{fig:toy} shows the representation space, including the population space and four different classes corresponding to classes with the minimum, mean, median and maximum area from among the 100 population classes that were sampled. As a comparison we also obtain the support of the population and the classes through the convex hull of the samples, even as this presents a number of practical challenges: (1) estimating convex hull in high-dimensions is computationally challenging, (2) convex hull overestimates the support due to outliers, and (3) cannot be easily adapted to obtain support as a function of desired FAR.

The capacity of the representation is now estimated as the ratio of the support area of the population and the class with median area, respectively. Table \ref{table:toy} shows the capacity estimates so obtained for this simplified representation space. Results on this example suggests that the ellipsoidal approximation of the representation is able to provide more accurate estimates of the capacity of the representation in comparison to the convex hull. Modeling the support of the representation through convex hulls is severely affected by outliers, resulting in an overestimate of the underlying support and area of the representation leading to overestimates of its capacity.

\subsection{Datasets}
\begin{figure*}[!ht]
    \centering
    \begin{subfigure}[t]{0.23\textwidth}
        \centering
        \includegraphics[width=0.23\textwidth]{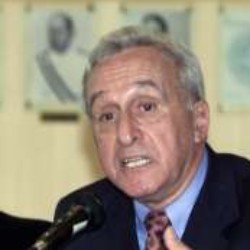}
        \includegraphics[width=0.23\textwidth]{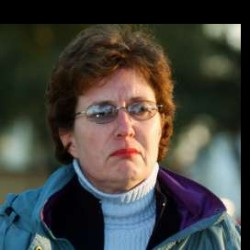}
        \includegraphics[width=0.23\textwidth]{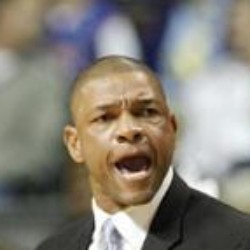}
        \includegraphics[width=0.23\textwidth]{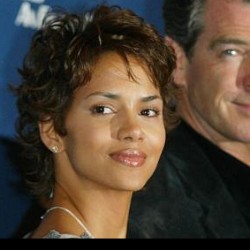}
        \includegraphics[width=0.23\textwidth]{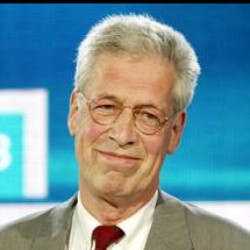}
        \includegraphics[width=0.23\textwidth]{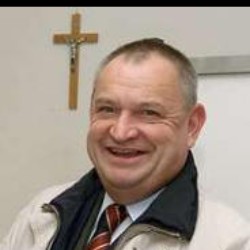}
        \includegraphics[width=0.23\textwidth]{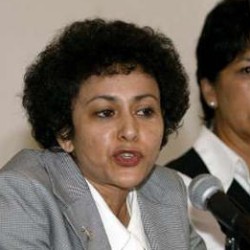}
        \includegraphics[width=0.23\textwidth]{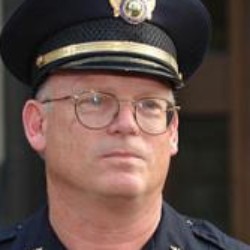}
        \includegraphics[width=0.23\textwidth]{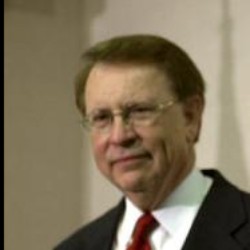}
        \includegraphics[width=0.23\textwidth]{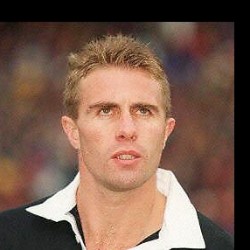}
        \includegraphics[width=0.23\textwidth]{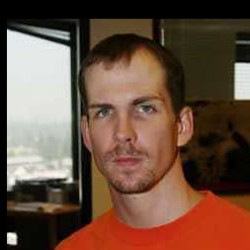}
        \includegraphics[width=0.23\textwidth]{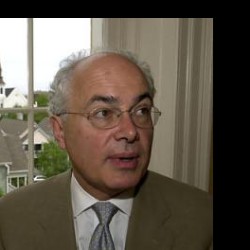}
        \caption{LFW \label{fig:datasets-lfw}}
    \end{subfigure}
    \hspace{3pt}
    \begin{subfigure}[t]{0.23\textwidth}
        \centering
        \includegraphics[width=0.23\textwidth]{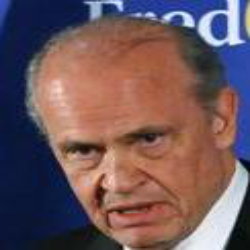}
        \includegraphics[width=0.23\textwidth]{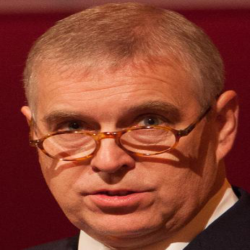}
        \includegraphics[width=0.23\textwidth]{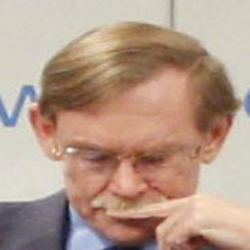}
        \includegraphics[width=0.23\textwidth]{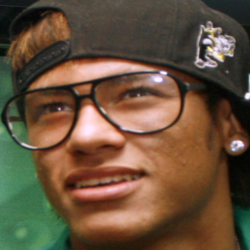}
        \includegraphics[width=0.23\textwidth]{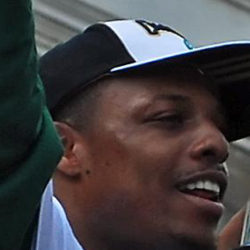}
        \includegraphics[width=0.23\textwidth]{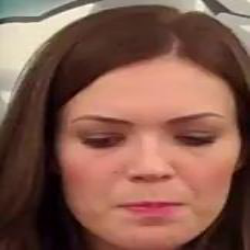}
        \includegraphics[width=0.23\textwidth]{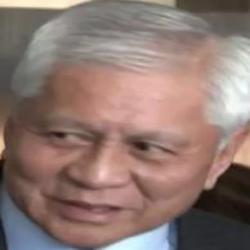}
        \includegraphics[width=0.23\textwidth]{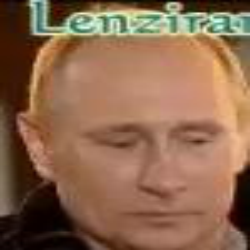}
        \includegraphics[width=0.23\textwidth]{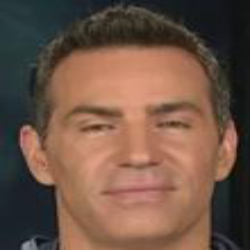}
        \includegraphics[width=0.23\textwidth]{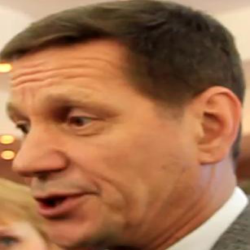}
        \includegraphics[width=0.23\textwidth]{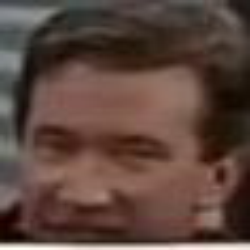}
        \includegraphics[width=0.23\textwidth]{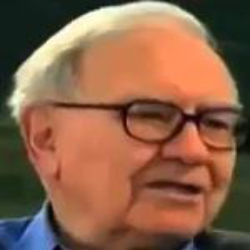}
        \caption{IJB-A \label{fig:datasets-ijba}}
    \end{subfigure}
    \hspace{3pt}
    \begin{subfigure}[t]{0.23\textwidth}
        \centering
        \includegraphics[width=0.23\textwidth]{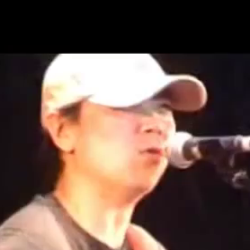}
        \includegraphics[width=0.23\textwidth]{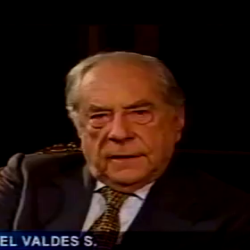}
        \includegraphics[width=0.23\textwidth]{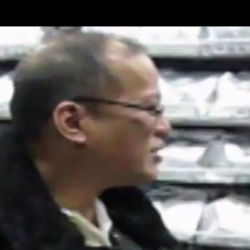}
        \includegraphics[width=0.23\textwidth]{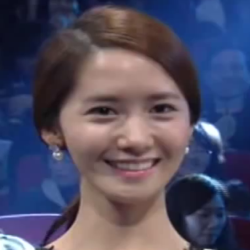}
        \includegraphics[width=0.23\textwidth]{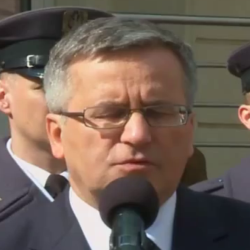}
        \includegraphics[width=0.23\textwidth]{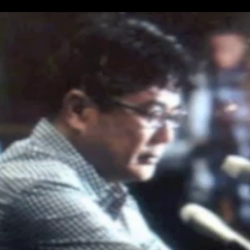}
        \includegraphics[width=0.23\textwidth]{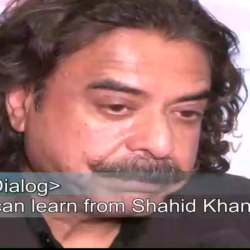}
        \includegraphics[width=0.23\textwidth]{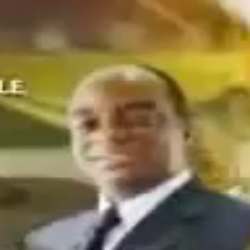}
        \includegraphics[width=0.23\textwidth]{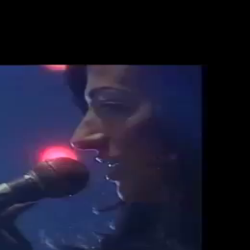}
        \includegraphics[width=0.23\textwidth]{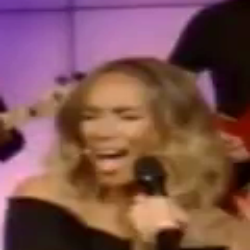}
        \includegraphics[width=0.23\textwidth]{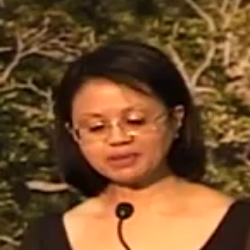}
        \includegraphics[width=0.23\textwidth]{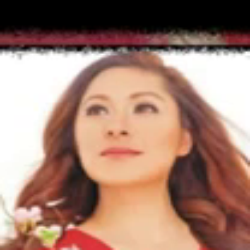}
        \caption{IJB-B \label{fig:datasets-ijbb}}
    \end{subfigure}
    \hspace{3pt}
    \begin{subfigure}[t]{0.23\textwidth}
        \centering
        \includegraphics[width=0.23\textwidth]{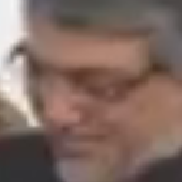}
        \includegraphics[width=0.23\textwidth]{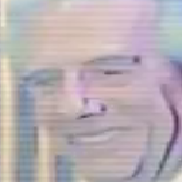}
        \includegraphics[width=0.23\textwidth]{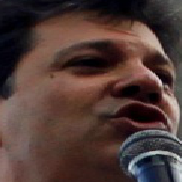}
        \includegraphics[width=0.23\textwidth]{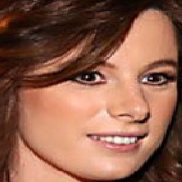}
        \includegraphics[width=0.23\textwidth]{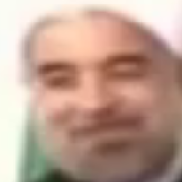}
        \includegraphics[width=0.23\textwidth]{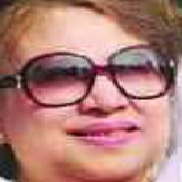}
        \includegraphics[width=0.23\textwidth]{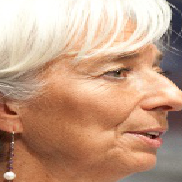}
        \includegraphics[width=0.23\textwidth]{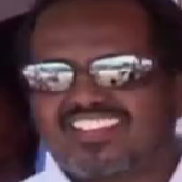}
        \includegraphics[width=0.23\textwidth]{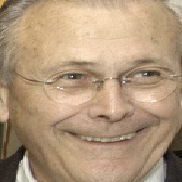}
        \includegraphics[width=0.23\textwidth]{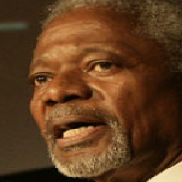}
        \includegraphics[width=0.23\textwidth]{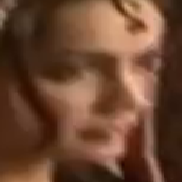}
        \includegraphics[width=0.23\textwidth]{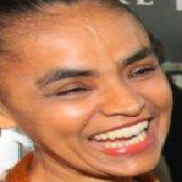}
        \caption{IJB-C \label{fig:datasets-ijbc}}
    \end{subfigure}
   	\caption{Example images from each of the four datasets for estimating the capacity shown in increasing order of complexity. Images from the CASIA dataset are not shown here because it is only used for training the student-teacher model.}
    \label{fig:datasets}
\end{figure*}

We utilize multiple large-scale face datasets, both for learning the \emph{teacher} and \emph{student} models as well as for estimating the capacity of the \emph{teacher}. Figure \ref{fig:datasets} shows a few examples of the kind of faces in each dataset.

\vspace{2pt}
\noindent\textbf{LFW \cite{huang2007labeled}:} 13,233 face images of 5,749 subjects, downloaded from the web. These images exhibit limited variations in pose, illumination, and expression, since only faces that could be detected by the Viola-Jones face detector \cite{viola2004robust} were included in the dataset. One limitation of this dataset is that only 1,680 subjects among the total of 5,749 subjects have more than one face image.

\vspace{2pt}
\noindent\textbf{CASIA \cite{yi2014learning}:} A large collection of labeled images downloaded from the web (based on names of famous personalities) typically used for training deep neural networks. It consists of 494,414 images across 10,575 subjects, with an average of about 500 face images per subject. This dataset is used for training both the \emph{teacher} and \emph{student} models.

\vspace{2pt}
\noindent\textbf{IJB-A \cite{klare2015pushing}:} IARPA Janus Benchmark-A (IJB-A) contains 500 subjects with a total of 25,813 images (5,399 still images and 20,414 video frames), an average of 51 images per subject. Compared to the LFW and CASIA datasets, the IJB-A dataset is more challenging due to: i) full pose variation making it difficult to detect all the faces using a commodity face detector, ii) a mix of images and videos, and iii) wider geographical variation of subjects. The face locations are provided with the IJB-A dataset (and used in our experiments when needed).

\vspace{2pt}
\noindent\textbf{IJB-B \cite{whitelam2017iarpa}:} IARPA Janus Benchmark-B (IJB-B) dataset is a superset of the IJB-A dataset consisting of 1,845 subjects with a total of 76,824 images (21,798 still images and 55,026 video frames from 7,011 videos), an average of 41 images per subject. Images in this dataset are labeled with ground truth bounding boxes and other covariate meta-data such as occlusions, facial hair and skin tone. A key motivation for the IJB-B dataset is to make the face database less constrained compared to the IJB-A dataset and have a more uniform geographic distribution of subjects across the globe.

\vspace{2pt}
\noindent\textbf{IJB-C \cite{maze2018iarpa}:} IARPA Janus Benchmark-C (IJB-C) dataset consists of 3,531 subjects with a total of 31,334 (21,294 face and 10,040 non-face) still images and 11,779 videos (117,542 frames), an average of 39 images per subject. This dataset emphasizes faces with full pose variations, occlusions and diversity of subject occupation and geographic origin. Images in this dataset are labeled with ground truth bounding boxes and other covariate meta-data such as occlusions, facial hair and skin tone.

\subsection{Face Representation Model}
We estimate the capacity of two different face representation models: (i) FaceNet introduced by Schroff \etal \cite{schroff2015facenet}, and (ii) SphereFace introduced by Liu \etal \cite{liu2017sphereface}. These models are illustrative of the state-of-the-art representations for face recognition.

The manifold projection and unfolding function is modeled as a multi-layer deep neural network with multiple residual \cite{He_2016_CVPR} modules consisting of fully-connected layers. Therefore, for a given image, the low-dimensional representation can be obtained by propagating the image through the \emph{original} face representation model and then through the manifold projection model. We refer to the combined model, i.e., \emph{original} representation and the projection model, as the \emph{teacher} model. Since the \emph{student} model is purposed to mimic the \emph{teacher} model, we base the \emph{student} network architecture on the \emph{teacher}'s\footnote{In the scenario where the \emph{teacher} is a black-box model, the design of the student network architecture needs more careful consideration but it also affords more flexibility. See Fig. \ref{fig:overview} for an illustration of this process.} architecture with a few notable exceptions. First, we introduce dropout before every convolutional layer of the network, including all the convolutional layers of the inception \cite{szegedy2015going} and residual \cite{He_2016_CVPR} modules and every linear layer of the manifold projection and unfolding modules. Second, the last layer of the network is modified to generate two outputs $\bm{\mu}$ and $\bm{\Sigma}$ instead of the output of the \emph{teacher} i.e., sample $\bm{y}$ of the noisy embedding.

\subsection{Face Recognition Performance}

Below we provide implementation details for learning the manifold projection and the \emph{student} networks. Subsequently, we demonstrate the ability of the \emph{student} model to maintain the discriminative performance of the \emph{original} models. 

\vspace{5pt}
\noindent \textbf{Implementation Details:} We use pre-trained models for both FaceNet\footnote{\url{https://github.com/davidsandberg/facenet}} and SphereFace\footnote{\url{https://github.com/wy1iu/sphereface}} as our \emph{original} face representation models. Before we extract features from these models, the face images are pre-processed and normalized to a canonical face image. The faces are detected and normalized using the joint face detection and alignment system introduced by Zhang \etal \cite{zhang2016joint}. Given the facial landmarks, the faces are normalized to a canonical image of size 182$\times$182 from which RGB patches of size 160$\times$160 are extracted as the input to the networks.

Given the features extracted from the \emph{original} representation, we train the manifold projection and unfolding networks on the CASIA WebFace dataset. The model is trained to minimize the multi-dimensional scaling loss function described in Eq. \ref{eq:deepmds} on randomly selected pairs of features vectors $\bm{x}_i$ and $\bm{x}_j$ from the dataset. Training is performed using the Adam \cite{kingma2014adam} optimizer with a learning rate of 3e-4 and the regularization parameter $\lambda=3\times10^{-4}$. We observed that using the cosine-annealing scheduler \cite{loshchilov2016sgdr} was critical to learning an effective mapping. We use a batch size of 256 image pairs and train the model for about 100 epochs.

The \emph{student} is trained to minimize the loss function defined in Eq. \ref{eq:all}, where the hyper-parameters are chosen through cross-validation. Training is performed through stochastic gradient descent with Nesterov Momentum 0.9 and weight decay 0.0005. We used a batch size of 64, a learning rate of 0.01 that is dropped by a factor of 2 every 20 epochs. We observed that it is sufficient to train the \emph{student} model for about 100 epochs for convergence. The \emph{student} model includes dropout with a probability of 0.05 after each convolutional layer and with a probability of 0.2 after each fully-connected layer in the manifold projection layers. At inference each image is passed through the \emph{student} network 1,000 times as a way of performing Monte-Carlo integration through the space of network parameters $\{\bm{w}_{\bm{\mu}},\bm{w}_{\bm{\Sigma}}\}$. These sampled outputs are used to empirically estimate the mean and covariance of the image embedding.

\vspace{5pt}
\noindent \textbf{Experiments:} We evaluate and compare the performance of the \emph{original} and \emph{student} models on the four test datasets, namely, LFW, IJB-A, IJB-B and IJB-C. To evaluate the \emph{student} model we estimate the face representation through Monte-Carlo integration. We pass each image through the \emph{student} model 1,000 times to extract $\{\bm{\mu}_i,\bm{\Sigma}_i\}_{i=1}^{1000}$ and compute $\bm{\mu}=\frac{1}{1000}\sum_{i=1}^{1000} \bm{\mu}_i$ as the representation. Following standard practice, we match a pair of representations through a nearest neighbor classifier i.e., by computing the euclidean distance $d_{ij} = \left\|\bm{y}_i-\bm{y}_j\right\|_2^2$ between the low-dimensional projected feature vectors $\bm{y}_i$ and $\bm{y}_j$.

We evaluate the face representation models on the LFW dataset using the BLUFR protocol \cite{liao2014benchmark} and follow the prescribed template based matching protocol, where each template is composed of possibly multiple images of the class, for the IJB-A, IJB-B and IJB-C datasets. Following the protocol in \cite{wang2015face}, we define the match score between templates as the average of the match scores between all pairs of images in the two templates.

\begin{figure*}
\centering
\begin{subfigure}[t]{0.245\textwidth}
    \centering
    \includegraphics[width=\textwidth]{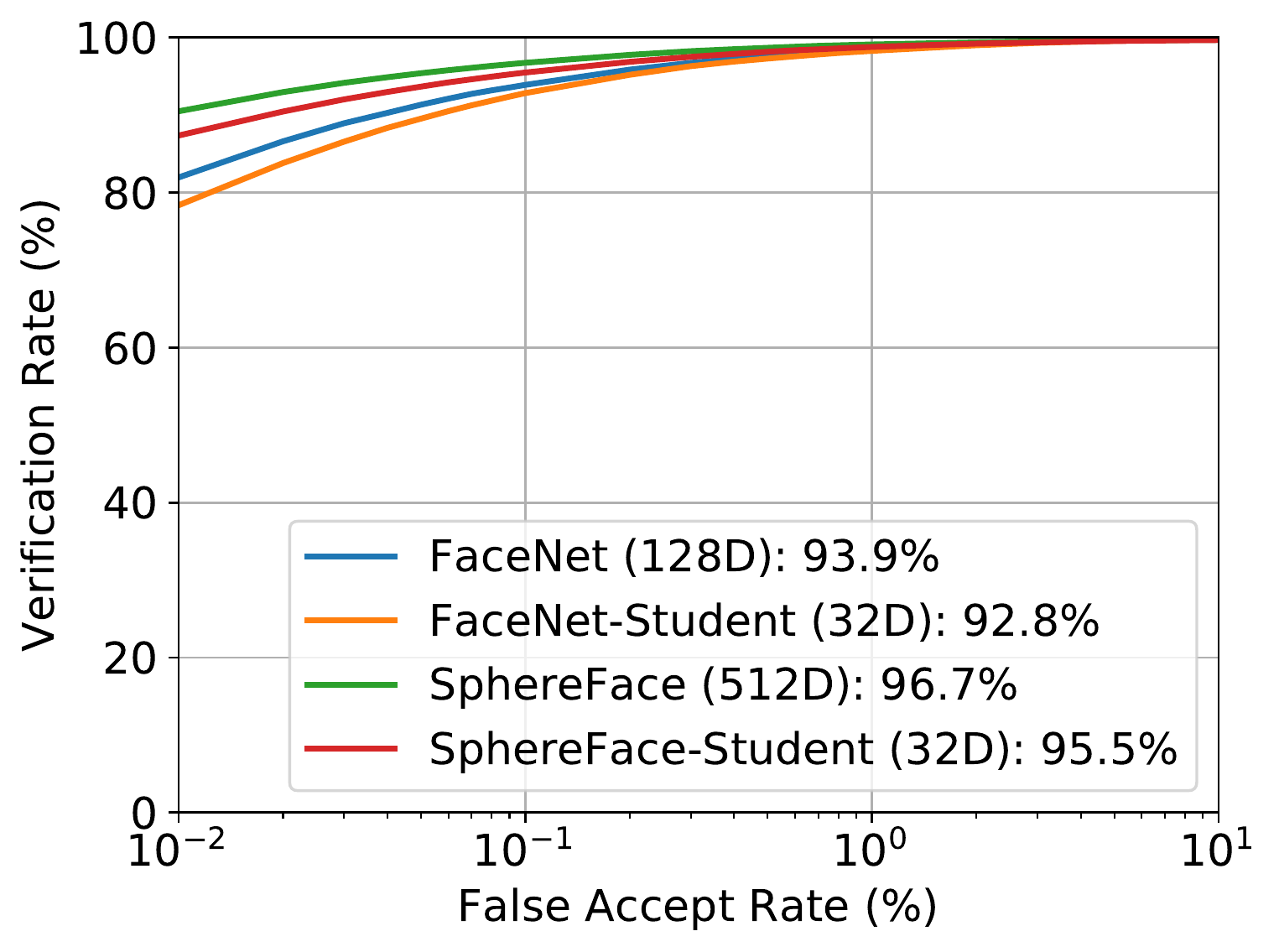}
    \caption{LFW}
    \label{fig:lfw-roc}
\end{subfigure}
\begin{subfigure}[t]{0.245\textwidth}
    \centering
    \includegraphics[width=\textwidth]{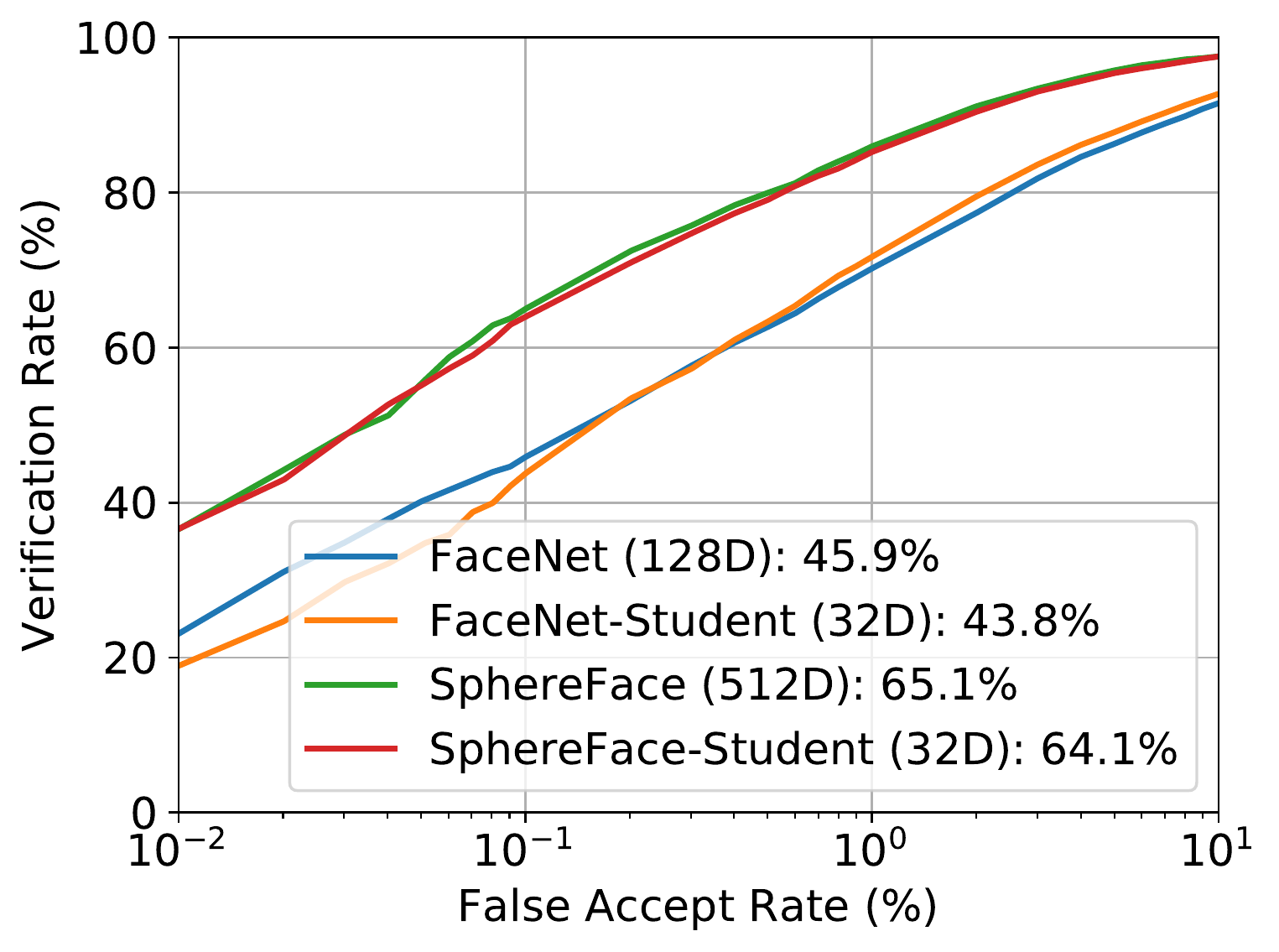}
    \caption{IJB-A}
    \label{fig:ijba-roc}
\end{subfigure}
\begin{subfigure}[t]{0.245\textwidth}
    \centering
    \includegraphics[width=\textwidth]{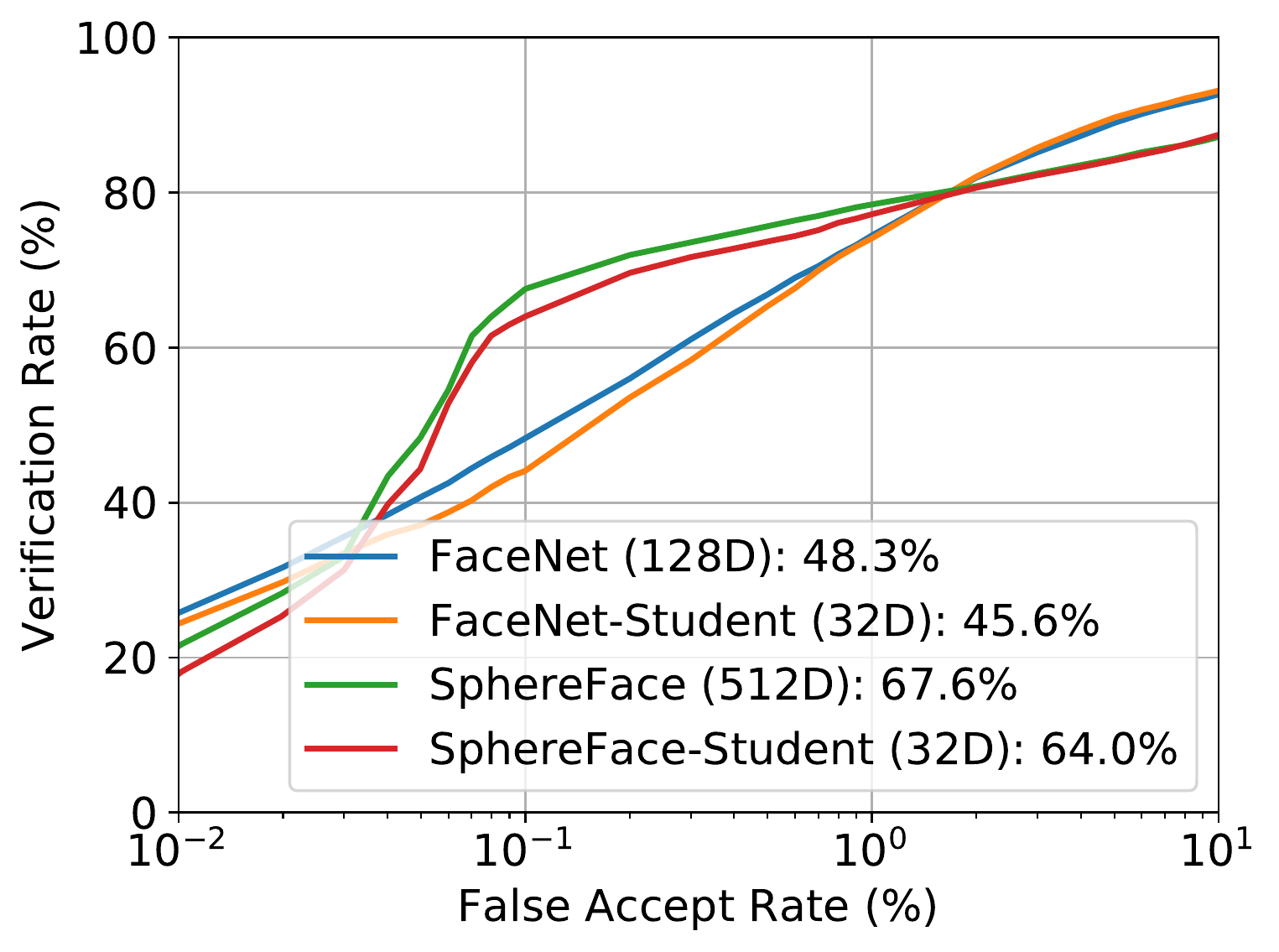}
    \caption{IJB-B}
    \label{fig:ijbb-roc}
\end{subfigure}
\begin{subfigure}[t]{0.245\textwidth}
    \centering
    \includegraphics[width=\textwidth]{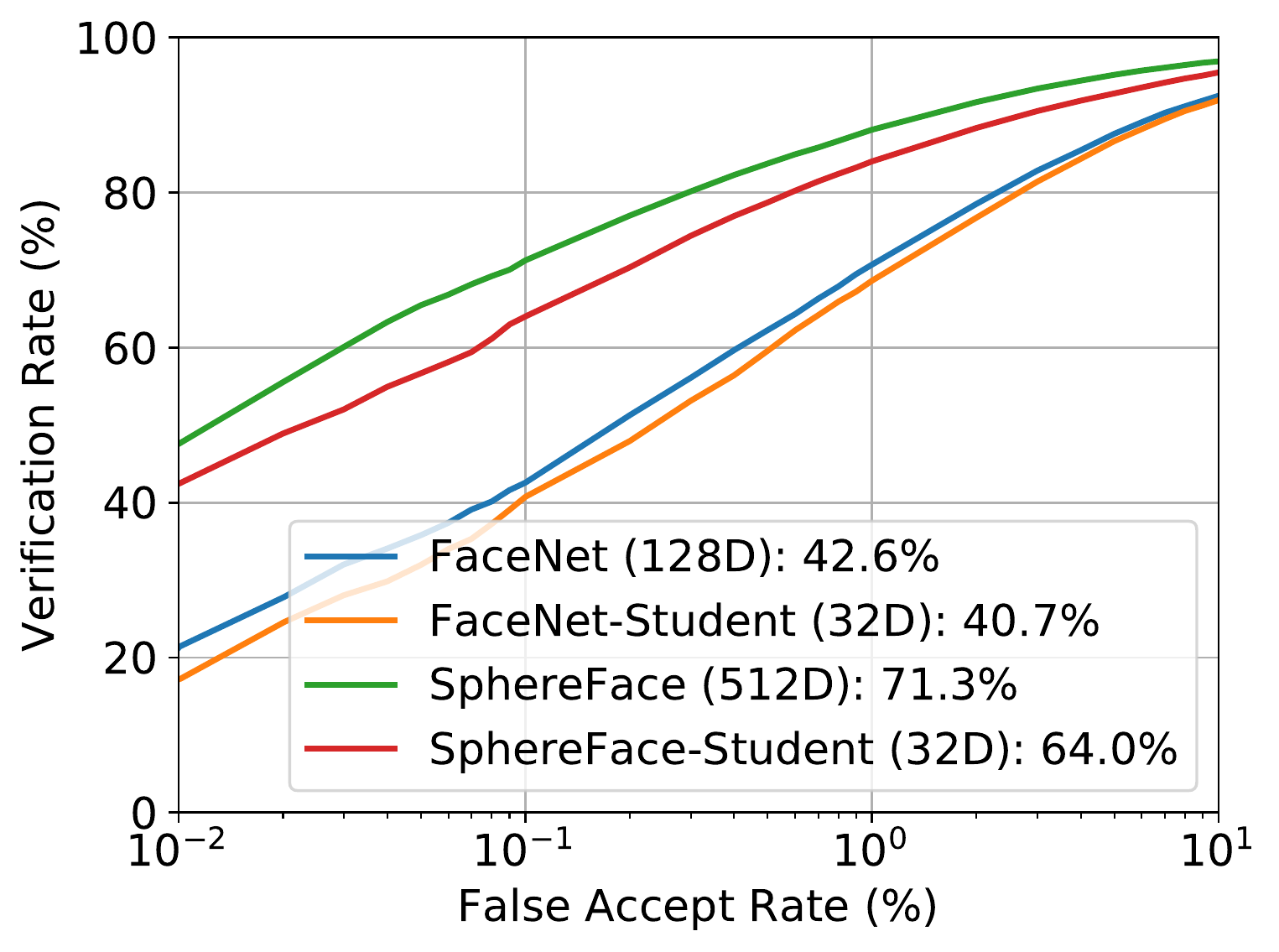}
    \caption{IJB-C}
    \label{fig:ijbc-roc}
\end{subfigure}
\caption{Face recognition performance of the \emph{original} and \emph{student} models on different datasets. We report the face verification performance of both FaceNet and SphereFace face representations, (a)  LFW evaluated through the BLUFR protocol, (b) IJB-A, (c) IJB-B, and (d) IJB-C evaluated through their respective matching protocol.}
\label{plot:roc}
\end{figure*}

Figure \ref{plot:roc} and Table \ref{table:recognition} report the performance of the \emph{original} and \emph{student} models, both FaceNet and SphereFace, on each of these datasets at different operating points. This comparison accounts for both the ability of the projection model to maintain the performance of the \emph{original} high-dimensional representation as well as the ability of the \emph{student} to mimic the \emph{teacher} while providing uncertainty estimates. We make the following observations: (1) The performance of DNN based representation on LFW, consisting largely of frontal face images with minimal pose variations and facial occlusions, is comparable to the state-of-the-art. However, its performance on IJB-A, IJB-B and IJB-C, datasets with large pose variations, is lower than state-of-the-art approaches. This is due to the template generation strategy that we employ and the fact that unlike these methods we do not fine-tune the DNN model on the IJB-A, IJB-B and IJB-C training sets. We reiterate that our goal in this paper is to estimate the capacity of a generic face representation as opposed to achieving the best verification performance on each individual datasets., and (2) Our results indicate that the \emph{student} models are able to mimic the \emph{teacher} models very well as demonstrated by the similarity of the receiving operating curves.

\begin{table*}[t]
    \centering
    \caption[Caption for LOF]{Face Recognition Results for FaceNet, SphereFace and State-of-the-Art\footnotemark\label{table:recognition}}
    \scalebox{0.9}{
    \begin{tabular}{lcccccccccccccc}
    \toprule
    \multirow{2}{*}{Dataset} & \multicolumn{2}{c}{\emph{Original}: FaceNet} && \multicolumn{2}{c}{\emph{Student}: FaceNet} && \multicolumn{2}{c}{\emph{Original}: SphereFace} && \multicolumn{2}{c}{\emph{Student}: SphereFace} && \multicolumn{2}{c}{State-of-the-Art}  \\
    \cline{2-3} \cline{5-6} \cline{8-9} \cline{11-12} \cline{14-15}
    & 0.1\% FAR & 1\% FAR && 0.1\% FAR & 1\% FAR && 0.1\% FAR & 1\% FAR && 0.1\% FAR & 1\% FAR && 0.1\% FAR & 1\% FAR \\
    \cline{2-3} \cline{5-6} \cline{8-9} \cline{11-12} \cline{14-15}
    LFW (BLUFR) & 93.90 & 98.51 && 92.83 & 98.28 && 96.74 & 99.11 && 95.49 & 98.79 && 98.88~\cite{wu2018light} & N/A\\
    IJB-A & 45.92 & 70.26 && 43.84 & 71.72 && 65.06 & 85.97 && 64.13 & 85.25 && 94.8 & 97.1~\cite{ranjan2018crystal}\\
    IJB-B & 48.31 & 74.47 && 45.56 & 74.10 && 67.58 & 80.81 && 64.02 & 80.63 && 93.7 & 97.5~\cite{xie2018comparator}\\
    IJB-C & 42.57 & 78.53 && 40.74 & 76.75 && 71.26 & 91.67 && 64.02 & 88.33 && 94.7 & 98.3~\cite{xie2018comparator}\\
    \bottomrule
    \end{tabular}}
\end{table*}
\footnotetext{The state-of-the-art face representation models are not available in the public domain.}

\subsection{Face Representation Capacity \label{sec:estimates}}
\begin{table}[!ht]
    \centering
    \caption{Capacity of Face Representation Model at 1\% FAR}
    \label{table:capacity}
    \scalebox{0.8}{
    \begin{tabular}{lccccccccccc}
    \toprule
    Dataset & Faces & FaceNet & SphereFace \\
    \midrule
    \multirow{4}{*}{LFW} & \multirow{4}{*}{
    \begin{minipage}{0.25\textwidth}
        \includegraphics[width=0.225\textwidth]{figs/lfw_samples/LFW_Almeida_Baptista_0001.jpg}
        \includegraphics[width=0.225\textwidth]{figs/lfw_samples/LFW_Claire_Leger_0001.jpg}
        \includegraphics[width=0.225\textwidth]{figs/lfw_samples/LFW_Doc_Rivers_0001.jpg}
        \includegraphics[width=0.225\textwidth]{figs/lfw_samples/LFW_Halle_Berry_0009.jpg}
    \end{minipage}
    } & \multirow{4}{*}{4.3$\times$$10^{6}$} & \multirow{4}{*}{2.6$\times$$10^{5}$} \\
    & & & & \\
    & & & & \\
    & & & & \\
    
    \midrule
    \multirow{4}{*}{IJB-A} & \multirow{4}{*}{
    \begin{minipage}{0.25\textwidth}
        \includegraphics[width=0.225\textwidth]{figs/ijba_samples/IJB_A_1196_0034.png}
        \includegraphics[width=0.225\textwidth]{figs/ijba_samples/IJB_A_4270_0128.png}
        \includegraphics[width=0.225\textwidth]{figs/ijba_samples/IJB_A_4336_0040.png}
        \includegraphics[width=0.225\textwidth]{figs/ijba_samples/IJB_A_5057_0137.png}
    \end{minipage}
    } & \multirow{4}{*}{6.3$\times$$10^{4}$} & \multirow{4}{*}{3.2$\times$$10^{6}$} \\
    & & & & \\
    & & & & \\
    & & & & \\

    \midrule
    \multirow{4}{*}{IJB-B} & \multirow{4}{*}{
    \begin{minipage}{0.25\textwidth}
        \includegraphics[width=0.225\textwidth]{figs/ijbb_samples/frames_7120.png}
        \includegraphics[width=0.225\textwidth]{figs/ijbb_samples/frames_18432.png}
        \includegraphics[width=0.225\textwidth]{figs/ijbb_samples/frames_20025.png}
        \includegraphics[width=0.225\textwidth]{figs/ijbb_samples/frames_25038.png}
    \end{minipage}
    } & \multirow{4}{*}{6.4$\times$$10^{4}$} & \multirow{4}{*}{2.4$\times$$10^{5}$} \\
    & & & & \\
    & & & & \\
    & & & & \\
    
    \midrule
    \multirow{4}{*}{IJB-C} & \multirow{4}{*}{
    \begin{minipage}{0.25\textwidth}
        \includegraphics[width=0.225\textwidth]{figs/ijbc_samples/9_frames_18263.png}
        \includegraphics[width=0.225\textwidth]{figs/ijbc_samples/377_img_3295.png}
        \includegraphics[width=0.225\textwidth]{figs/ijbc_samples/39_frames_11816.png}
        \includegraphics[width=0.225\textwidth]{figs/ijbc_samples/399_img_4745.png}
    \end{minipage}
    } & \multirow{4}{*}{2.7$\times$$10^{4}$} & \multirow{4}{*}{8.4$\times$$10^{4}$} \\
    & & & & \\
    & & & & \\
    & & & & \\

    \bottomrule
    \end{tabular}}
\end{table}

Having demonstrated the ability of the \emph{student} model to be an effective proxy for the \emph{original} representation manifold, we indirectly estimate the capacity of the \emph{original} model by estimating the capacity of the \emph{student} model.

\vspace{2pt}
\noindent \textbf{Implementation Details:} We estimate the capacity of the face representations by evaluating Eq. \ref{eq:volume}. For each of the datasets we empirically determine the shape and size of the population hyper-ellipsoid $\bm{\Sigma}_{\bm{y}_c}$ and the class-specific hyper-ellipsoids $\bm{\Sigma}_{\bm{z}_c}$. These quantities are computed through the predictions obtained by sampling the weights ($\bm{w}_{\bm{\mu}},\bm{w}_{\bm{\Sigma}}$) of the model, via dropout. We obtain 1,000 such predictions for a given image, by feeding the image through the \emph{student} network a 1,000 different times with dropout. For robustness against outliers we only consider classes with at least two images per class for LFW and five images per class for all the other datasets for the capacity estimates.

\begin{figure*}[!ht]
    \centering
    \begin{subfigure}[t]{0.31\textwidth}
        \centering
        \includegraphics[width=\textwidth]{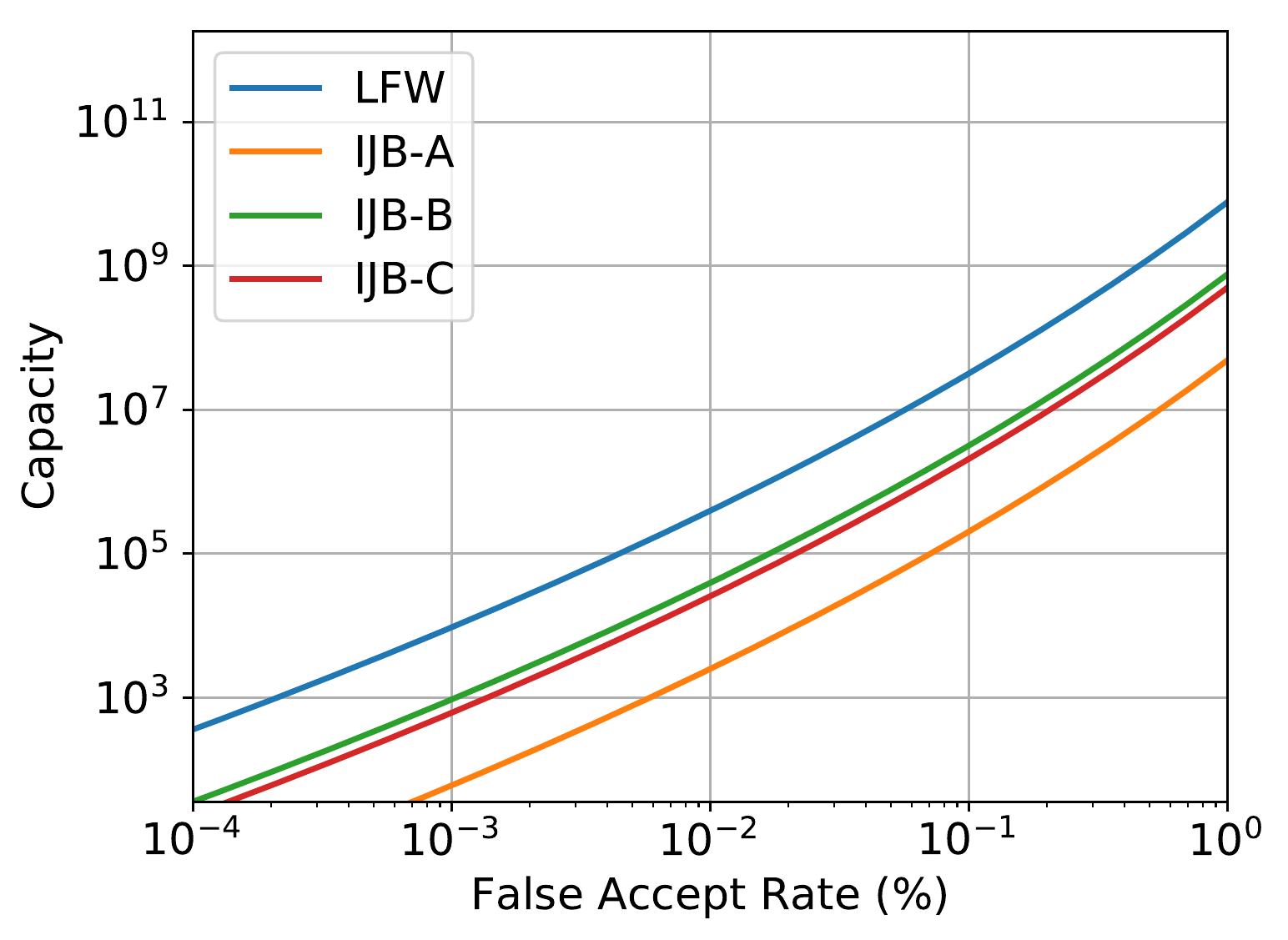}
        \caption{FaceNet \cite{schroff2015facenet}\label{plot:facenet-capacity}}
    \end{subfigure}
    \begin{subfigure}[t]{0.31\textwidth}
        \centering
        \includegraphics[width=\textwidth]{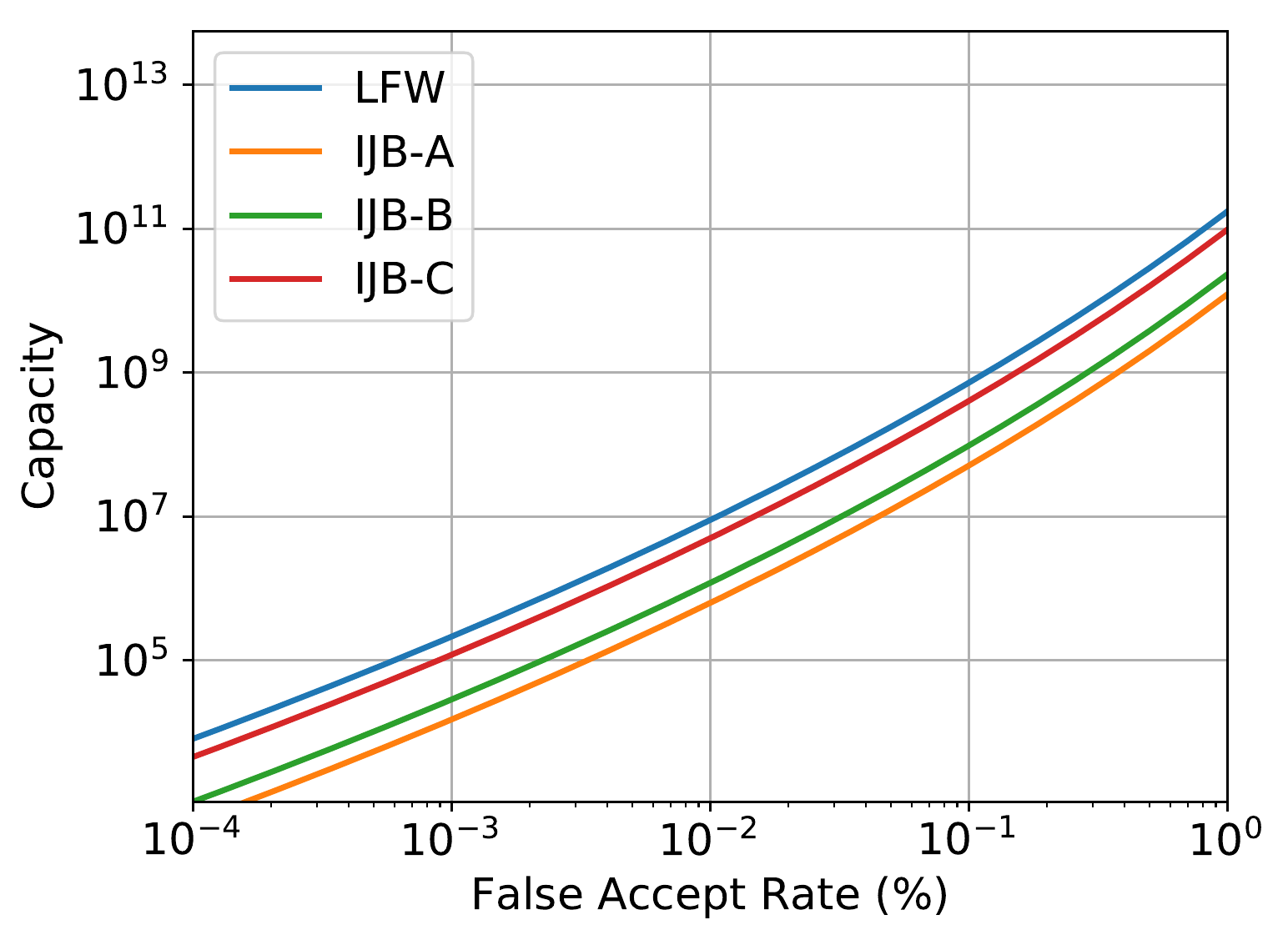}
        \caption{SphereFace \cite{liu2017sphereface}\label{plot:sphereface-capacity}}
    \end{subfigure}
    \begin{subfigure}[t]{0.32\textwidth}
        \centering
        \includegraphics[width=\textwidth]{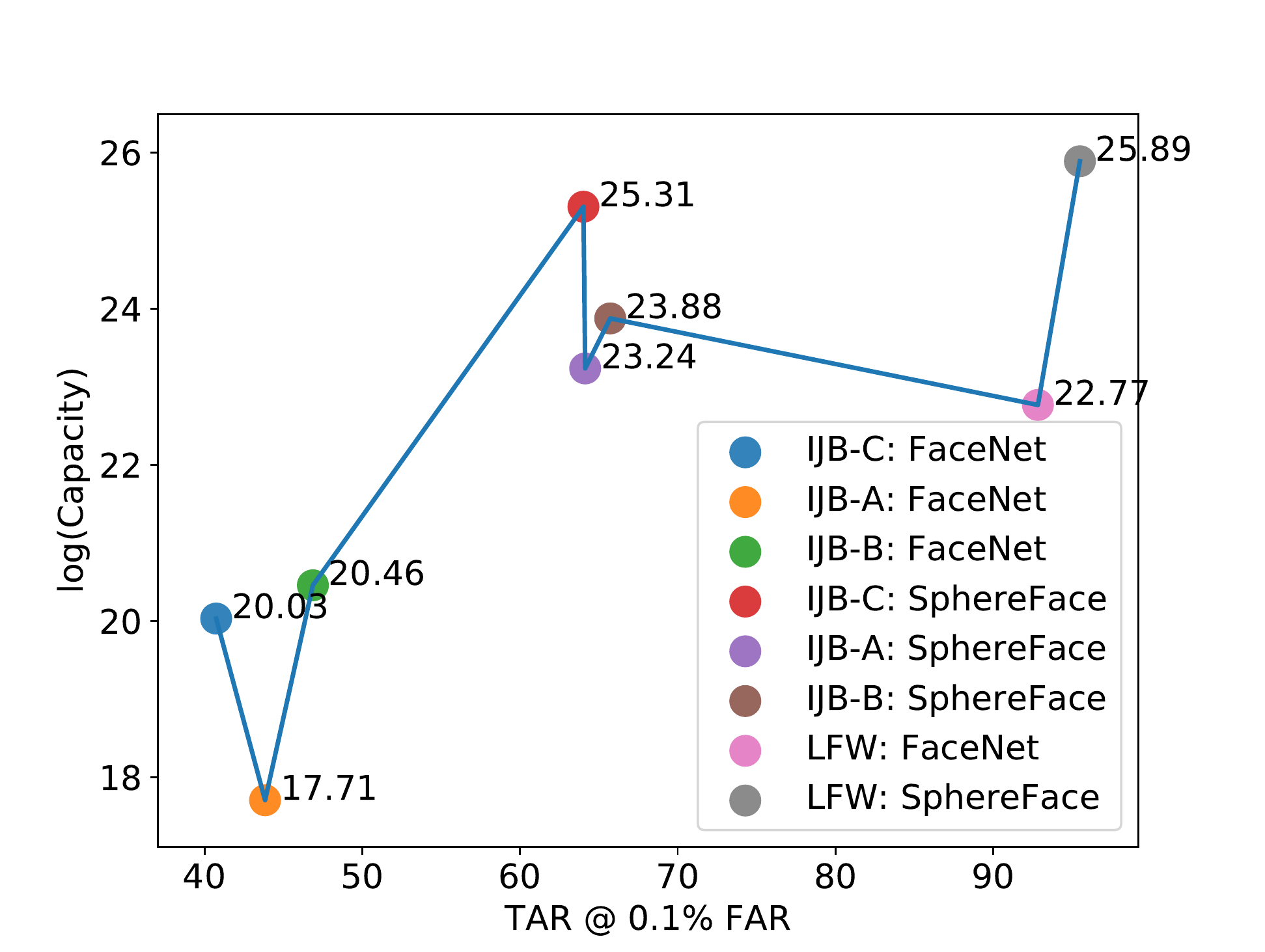}
        \caption{Capacity vs TAR\label{plot:capcity_with_tar}}
    \end{subfigure}
    \caption{Capacity estimates across different datasets for the (a) FaceNet \cite{schroff2015facenet} and (b) SphereFace \cite{liu2017sphereface} representations as function of different false accept rates. Under the limit, the capacity tends to zero as the FAR tends to zero. Similarly, the capacity tends to $\infty$ as the FAR tends to 1.0. (c) Logarithmic values of capacity on different datasets versus the corresponding TAR @ 0.1\% FAR. \label{plot:capacity}}
\end{figure*}

\vspace{5pt}
\noindent \textbf{Capacity Estimates:} Table \ref{table:capacity} reports the capacity of DNN based face representations estimated on different datasets at 1\% FAR (i.e., when $r_{\bm{y}_c}=r_{\bm{z}_c}$). We make the following observations from our numerical results: The upper bound on the capacity estimate of the FaceNet and SphereFace models in constrained scenarios (LFW) is of the order of $\approx 10^{6}$, in unconstrained environments (IJB-A, IJB-B and IJB-C) is of the order of $\approx 10^{5}$ under the general model of a hyper-ellipsoid with the class corresponding to maximum noise. Therefore, theoretically, the representation should be able to resolve $10^{6}$ and $10^{5}$ subjects with a true acceptance rate (TAR) of 100\% at a FAR of 1\% under the constrained and unconstrained operational settings, respectively. While this capacity estimate is on the order of the population of a large city, in practice, the performance of the representation is lower than the theoretical performance, about 95\% across only 10,000 subjects in the constrained and only 50\% across 3,531 subjects in the unconstrained scenarios. These results suggest that our capacity estimates are an upper bound on the actual performance of face recognition systems in practice, especially under unconstrained scenarios. The relative order of the capacity estimates, however, mimics the relative order of the verification accuracy on these datasets as shown in Fig. \ref{plot:capcity_with_tar}.

We extend the capacity estimates presented above to establish capacity as a function of different operating points, as defined by different false accept rates. We define $r_{\bm{y}_c}$ and $r_{\bm{z}_c}$ corresponding to the desired operating points and evaluate Eq. \ref{eq:volume}. In all our experiments we choose $r_{\bm{y}_c}$ to encompass 99\% of the classes within the population hyper-ellipsoid. Different FARs define different decision boundary contours that, in turn, define the size of the class-specific hyper-ellipsoid. Figures \ref{plot:facenet-capacity} and \ref{plot:sphereface-capacity} shows how the capacity of the representation changes as a function of the FARs for different datasets. We note that at the operating point of $FAR=0.1\%$, the capacity of the maximum face representation is $\approx 10^{5}$ in the constrained and $\approx 10^{3}$ in the unconstrained case. However, at stricter operating points (FAR of $0.001\%$ or $0.0001\%$), that is more meaningful at larger scales of operation \cite{kemelmacher2016megaface}, the capacity of the FaceNet representation is significantly lower (63 and 6, respectively for IJB-C) than the typical desired scale of operation of face recognition systems. These results suggest a significant room for improvement in face representation.

\subsection{Ablation Studies}

\vspace{2pt}
\noindent \textbf{DNN and PCA:} We seek to compare the capacity of classical PCA based EigenFaces \cite{turk1991face} representation of image pixels and the DNN based representation. These are illustrative of the two extremes of various face representations proposed in the literature with FaceNet and SphereFace providing close to state-of-the-art recognition performance. The FaceNet and SphereFace representations are based on non-linear multi-layered deep convolutional network architectures. EigenFaces, in contrast, is a linear model for representing faces. The capacity of Eigenfaces is $\approx 10^{0}$, which is significantly lower than the capacity of DNN based representations. Eigenfaces, by virtue of being based on linear projections of the raw pixel values, is unable to scale beyond a handful of identities, while the DNN representations are able to resolve significantly more number of identities. The relative difference in the capacity is also reflected in the vast difference in the verification performance between the two representations.
\begin{table}
    \centering
    \caption{\footnotesize IJB-C Capacity at 1\% FAR Across Intra-Class Uncertainty \label{table:classes}}
    \scalebox{0.8}{
    \begin{tabular}{ccccccccc}
        \toprule
        Model && Min && Mean && Median && Max \\
        \cline{1-1} \cline{3-3} \cline{5-5} \cline{7-7} \cline{9-9}
        && && && && \\
        FaceNet && $1.6\times10^{14}$ && $6\times10^8$ && $5.0\times10^8$ && $2.7\times10^4$ \\
        SphereFace && $4.9\times10^{16}$ && $1.1\times10^{11}$ && $9.8\times10^{10}$ && $8.4\times10^4$ \\
        \bottomrule
    \end{tabular}}
\end{table}

\begin{figure*}[!ht]
    \centering
    \begin{subfigure}[t]{0.32\textwidth}
        \centering
        \includegraphics[height=0.05\textheight]{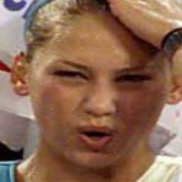}
        \includegraphics[height=0.05\textheight]{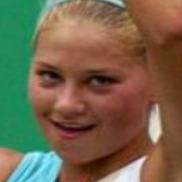}
        \includegraphics[height=0.05\textheight]{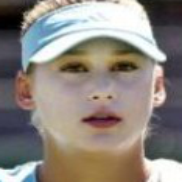}
        \includegraphics[height=0.05\textheight]{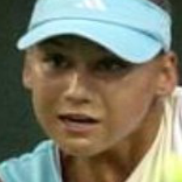}
        \includegraphics[height=0.05\textheight]{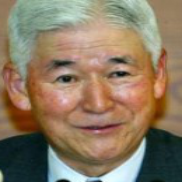}
        \includegraphics[height=0.05\textheight]{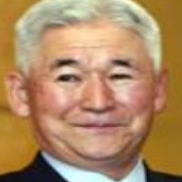}
        \includegraphics[height=0.05\textheight]{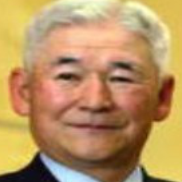}
        \includegraphics[height=0.05\textheight]{figs/noise/DNN/lfw/small/2.png}
        \caption{LFW}
    \end{subfigure}
    \begin{subfigure}[t]{0.3\textwidth}
        \centering
        \includegraphics[height=0.05\textheight]{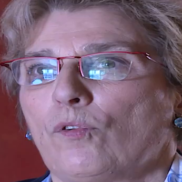}
        \includegraphics[height=0.05\textheight]{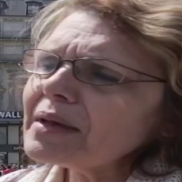}
        \includegraphics[height=0.05\textheight]{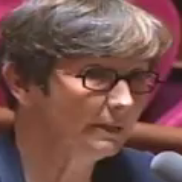}
        \includegraphics[height=0.05\textheight]{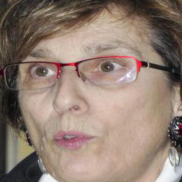}
        \includegraphics[height=0.05\textheight]{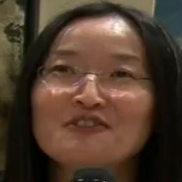}
        \includegraphics[height=0.05\textheight]{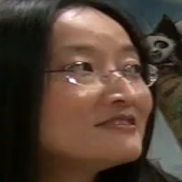}
        \includegraphics[height=0.05\textheight]{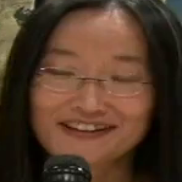}
        \includegraphics[height=0.05\textheight]{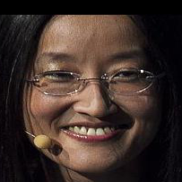}
        \caption{IJB-B}
    \end{subfigure}
    \begin{subfigure}[t]{0.37\textwidth}
        \centering
        \includegraphics[height=0.05\textheight]{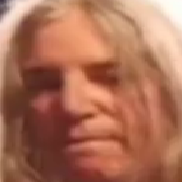}
        \includegraphics[height=0.05\textheight]{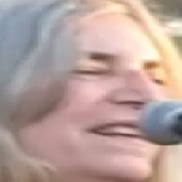}
        \includegraphics[height=0.05\textheight]{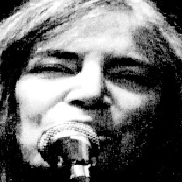}
        \includegraphics[height=0.05\textheight]{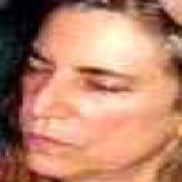}
        \includegraphics[height=0.05\textheight]{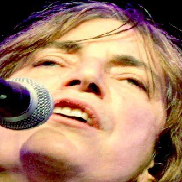}
        \includegraphics[height=0.05\textheight]{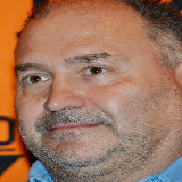}
        \includegraphics[height=0.05\textheight]{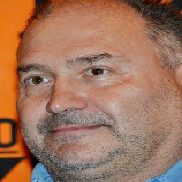}
        \includegraphics[height=0.05\textheight]{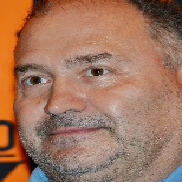}
        \includegraphics[height=0.05\textheight]{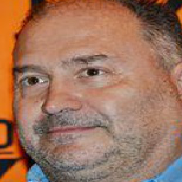}
        \includegraphics[height=0.05\textheight]{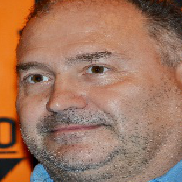}
        \caption{IJB-C}
    \end{subfigure}
    \caption{Example images of classes that correspond to different sizes of the class-specific hyper-ellipsoids, based on the SphereFace representation, for different datasets considered in the paper. \textbf{Top Row:} Images of the class with the largest class-specific hyper-ellipsoid for each database. Notice that in the case of a database with predominantly frontal faces (LFW), large variations in facial appearance lead to the greatest uncertainty in the class representation. On more challenging datasets (IJB-B, IJB-C), the face representation exhibits most uncertainty due to pose variations. \textbf{Bottom Row:}  Images of the class with the smallest class-specific hyper-ellipsoid for each database. As expected, across all the datasets, frontal face images with the minimal change in appearance result in the least amount of uncertainty in the class representation.}
    \label{fig:class-variability-images}
\end{figure*}

\vspace{2pt}
\noindent \textbf{Data Bias:} Our capacity estimates are critically dependent on the representational support of the canonical class. In other words, the capacity expression in Eq. \ref{eq:volume} depends on $\bm{\Sigma}_{\bm{z}_c}$, that is representative of the demographics and intra-class variability of the subjects in the population of interest. However, the hyper-ellipsoids corresponding to various classes could potentially be of a different size. For instance, in Fig. \ref{fig:embedding} each class-specific manifold is of different sizes, orientation and shape. Precisely defining or identifying a canonical subject, from among all possible identities, is in itself a challenging task and beyond the scope of this paper. In Table \ref{table:classes} we report the capacity for different choices of classes (subjects) from the IJB-C dataset i.e., classes with the minimum, mean, median and maximum hyper-ellipsoid volume, thereby ranging from classes with very low intra-class variability and classes with very high intra-class variability. Datasets whose class distribution is similar to the distribution of the data that was used to train the face representation, are expected to exhibit low intra-class uncertainty, while datasets with classes that are out of the training distribution can potentially have high intra-class uncertainty, and consequently lower capacity. Figure \ref{fig:class-variability-images} show examples of the images corresponding to the lowest and highest intra-class variability in each dataset.

Empirically, we observed that classes with the smallest hyper-ellipsoid are typically classes with very few images and very little variation in facial appearance. Similarly, classes with high intra-class uncertainty are typically classes with a very large number of images spanning a wide range of variations in pose, expression, illumination conditions etc., variations that one can expect under any real-world deployments of face recognition systems. Coupled with the fact that the capacity of the face representation is estimated from a very small sample of the population (less than 11,000 subjects), we argue that the class with large intra-class uncertainty within the datasets considered in this paper is a reasonable proxy of a canonical subject in unconstrained real-world deployments of face recognition systems.

\begin{table}
    \centering
    \caption{\footnotesize IJB-C Capacity at 1\% FAR Across Manifold Support \label{table:gaussian}}
    \scalebox{0.8}{
    \begin{tabular}{ccccccccc}
        \toprule
        \multirow{2}{*}{Model} && \multirow{2}{*}{Hypersphere} && Hyper-Ellipsoid && \multirow{2}{*}{Hyper-Ellipsoid} \\
        && && (Axis-Aligned) && \\
        \cline{1-1} \cline{3-3} \cline{5-5} \cline{7-7}
        && && && && \\
        FaceNet && $1.5\times10^{3}$ && $9.2\times10^2$ && $2.7\times10^4$ \\
        SphereFace && $6.7\times10^{3}$ && $7.2\times10^{3}$ && $8.4\times10^4$ \\
        \bottomrule
    \end{tabular}}
\end{table}
\vspace{2pt}
\noindent \textbf{Gaussian Distribution Parameterization:} For the sake of efficiency we made the same modeling assumption for both the global shape of the embedding and the embedding shape of each class. The capacity estimates obtained thus far are by modeling the manifolds as unconstrained hyper-ellipsoids. We now obtain capacity estimates for different modeling assumptions on the shape of these entities. For instance the shapes could also be modeled as hyper-spheres corresponding to a diagonal covariance matrix with the same variance in each dimension. We generalize the hyper-sphere model to an axis aligned hyper-ellipsoid corresponding to a diagonal covariance matrix with possibly different variances along each dimension. Table \ref{table:gaussian}  shows the capacity estimates on the IJB-C dataset at 1\% FAR. We observe that the capacity estimates of the anisotropic Gaussian (hyper-ellipsoid) are two orders of magnitude higher than the capacity estimates of the reduced approximations, hyper-sphere (isotropic Gaussian) and axis-aligned hyper-ellipsoid. At the same time, the isotropic and the axis-aligned hyper-ellipsoid approximations result in very similar capacity estimates.


\section{Conclusion}
Face recognition is based on two underlying premises: persistence (invariance of face representation over time) and capacity (number of distinct identities a face representation ca resolve). While face longitudinal studies \cite{best2018longitudinal} have addressed the persistence property, very little attention has been devoted to the capacity problem that is addressed here. The face representation process was modeled as a low-dimensional manifold embedded in high-dimensional space. We estimated the capacity of a face representation as a ratio of the volume of the population and class-specific manifolds as a function of the desired false acceptance rate. Empirically, we estimated the capacity of two deep neural network based face representations: FaceNet and SphereFace. Numerical results yielded a capacity of $10^{5}$ at a FAR of 1\%. At lower FAR of $0.001\%$, the capacity dropped-off significantly to only 70 under unconstrained scenarios, impairing the scalability of the face representation. There does exist a large gap between the theoretical and empirical verification performance of the representations indicating that there is a significant scope for improvement in the discriminative capabilities of current state-of-the-art face representations.

As face recognition technology makes rapid strides in performance and witnesses wider adoption, quantifying the capacity of a given face representation is an important problem, both from an analytical as well as from a practical perspective. However, due to the challenging nature of finding a closed-form expression of the capacity, we make simplifying assumptions on the distribution of the population and specific classes in the representation space. Our experimental results demonstrate that even this simplified model is able to provide reasonable capacity estimates of a DNN based face representation. Relaxing the assumptions of the approach presented here is an exciting direction of future work, leading to more realistic capacity estimates.

\ifCLASSOPTIONcaptionsoff
  \newpage
\fi

\bibliographystyle{IEEEtran}
\bibliography{mybib}
\end{document}